\documentclass[10pt,twocolumn,letterpaper]{article}

\usepackage[datasets]{wacv}
\usepackage{graphicx}
\usepackage{amsmath}
\usepackage{amssymb}
\usepackage{booktabs}
\usepackage{tikz}
\usetikzlibrary{arrows.meta,positioning,shapes.geometric}
\usepackage[numbers,sort&compress]{natbib}
\usepackage{multirow}
\usepackage{xcolor}
\usepackage{xspace}
\usepackage{algorithm}
\usepackage{algorithmic}
\usepackage{tabularx}
\usepackage[table]{xcolor}
\usepackage{array}

\usepackage[accsupp]{axessibility}  

\usepackage[pagebackref,breaklinks,colorlinks]{hyperref}

\usepackage{microtype}

\newcommand{\ourmodel}{\textsc{M3Track}\xspace}
\newcommand{\ourdata}{\textsc{Refer-CityFlow}\xspace}
\newif\ifreviewhighlight
\reviewhighlightfalse
\newcommand{\reviewedit}[1]{\ifreviewhighlight \textcolor{blue}{#1}\else #1\fi}

\definecolor{myblue}{HTML}{185FA5}
\definecolor{myorange}{HTML}{BA7517}
\definecolor{myred}{HTML}{A32D2D}

\newcommand{\cc}[1]{}                          

\definecolor{rbteal}{rgb}{0.0,0.42,0.45}
\ifdefined\showmarks

\else

\fi

\newcommand{\newcc}[1]{}
\newcommand{\eli}[1]{}
\newcommand{\eric}[1]{}
\newcommand{\hao}[1]{}
\newcommand{\nam}[1]{}
\newcommand{\ngan}[1]{}
\newcommand{\tam}[1]{}

\def\wacvPaperID{2943} 
\def\confName{WACV}
\def\confYear{2027}

\begin{document}

\title{Refer-CityFlow and M3Track: A Benchmark Dataset and Reference Framework for Language-Driven Multi-Target Multi-Camera Tracking}

\author{
Phuong T. Nguyen$^{1}$\thanks{Equal contribution.}\quad
Hao Vo$^{2}$\footnotemark[1]\quad
Tam Bang$^{1}$\footnotemark[1]\quad
Hoang K. Tran$^{1}$\footnotemark[1]\\
Hoang-Minh Dau$^{3}$\quad
Mahshid Malazizi$^{1}$\quad
Tuan-Anh Hoang$^{3}$\quad
Mina Sartipi$^{1}$\\
Thanh-Nam Doan$^{4}$\quad
Ngan Le$^{2}$\quad
Hoang H. Nguyen$^{1}$\\[4pt]
$^{1}$University of Tennessee at Chattanooga \quad
$^{2}$University of Arkansas\\
$^{3}$RMIT International University Vietnam \quad
$^{4}$Independent Researcher\\
{\tt\small qlg645@mocs.utc.edu, huuhoang-nguyen@utc.edu}
}
\maketitle

\begin{abstract}
Referring multi-object tracking enables users to specify targets through natural
language, but existing Refer-MOT benchmarks primarily focus on single-camera
settings. We formalize \textbf{Referring Multi-Target Multi-Camera Tracking
(Refer-MTMCT)}, where free-form queries specify identity-consistent trajectories
to be recovered across a network of sequential, sparse-overlap cameras. We introduce \textbf{\ourdata}, a benchmark built on CityFlow's global cross-camera identity annotations and augmented with language queries
requiring joint language grounding, cross-camera association, and
spatio-temporal reasoning. \ourdata covers 96 videos from 40 roadside cameras
with 660+ identity-consistent cross-camera trajectories, 4{,}362 trajectory-level
descriptions, and 7{,}699 queries, including 1{,}699 real-user queries collected
from 50 participants.
\reviewedit{Reference experiments on the four-camera Scenario~02 show that single-camera referring trackers
reach 14.95--18.28 HOTA on real-user queries, while the evaluated
cross-view method achieves about 0.05 HOTA in direct transfer.}
To establish reference performance, we provide \textbf{\ourmodel}, a modular multimodal reference framework in which a large language model (LLM)-based agent orchestrates interchangeable detection, tracking, and
cross-camera association modules. \ourmodel reaches 44.10 HOTA and 65.40 IDF1 on the same real-user query set. 
The project repository is available at
\url{https://github.com/phuongnguyen194/Refer-CityFlow-and--M3Track}.

\end{abstract}

\section{Introduction}
\label{sec:intro}

\begin{figure*}[t]
  \centering
  \begin{minipage}{\textwidth}
    \centering
    \begin{minipage}{0.20\textwidth}
      \centering\footnotesize
      \fcolorbox{blue!40}{blue!8}{\parbox{0.92\linewidth}{\textbf{Query:} ``Track the
      black SUV from \emph{camera~6}, reappearing in \emph{cameras~7 and~9}.''}}
    \end{minipage}\hfill
    \begin{minipage}{0.78\textwidth}
      \centering
      \includegraphics[width=\linewidth]{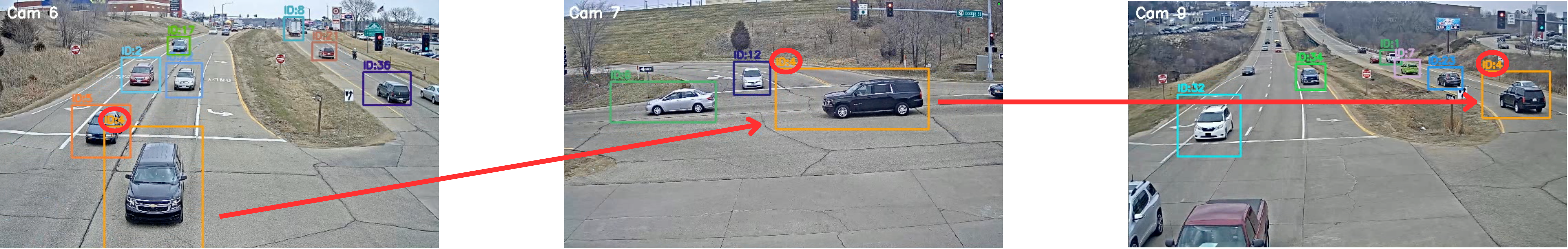}\\[1pt]
      {\scriptsize\itshape One query $\to$ one identity (GID~4) tracked across
      \emph{sequential} cameras Cam~6 $\to$ 7 $\to$ 9.}
    \end{minipage}
  \end{minipage}

  \vspace{4pt}
  \hrule
  \vspace{4pt}

  \begin{minipage}{\textwidth}
    \centering
    \includegraphics[
        width=\linewidth,
        keepaspectratio
    ]{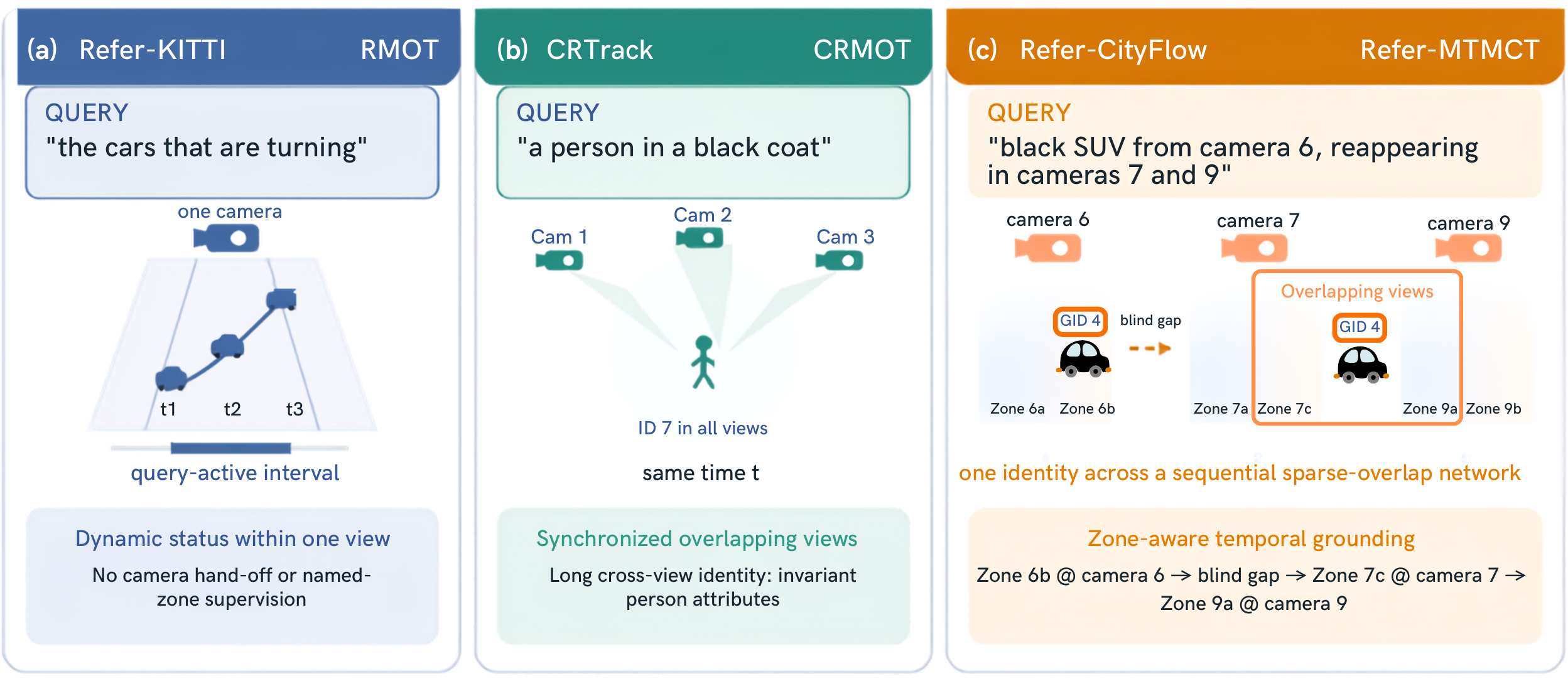}
  \end{minipage}
    \caption{\textbf{\ourdata and the Refer-MTMCT setting.}
    \emph{Top}: one free-form query resolves to an identity-consistent trajectory
    across sequential cameras (Cam~6$\to$7$\to$9).
    \emph{Bottom}: comparison with prior referring settings:
    \textbf{(a)} RMOT operates within one camera;
    \textbf{(b)} CRMOT~\cite{chen2025cross} uses synchronized overlapping views of
    one scene; and
    \textbf{(c)} Refer-MTMCT spans a sequential, sparse-overlap camera network,
    where the query defines the target set and combines appearance, cross-camera,
    and spatio-temporal constraints.}
  \label{fig:intro}
\end{figure*}

Referring multi-object tracking (Refer-MOT)~\citep{wu2023referring,
nguyen2024type, du2024ikun, tran2024z, anh2024tp} lets users specify targets
through natural language rather than visual initialization. Existing Refer-MOT benchmarks,
however, largely assume a single camera, while real camera networks often
require following a described target across a camera network, e.g.,
the query ``the black SUV from camera~6, reappearing in cameras~7 and~9''
requires both language grounding and identity preservation across views.

Existing benchmarks cover only parts of this problem
(Fig.~\ref{fig:intro}). Refer-MOT datasets~\citep{wu2023referring,
zhang2024bootstrapping,chen2025referbdd} provide rich language but remain
single-camera. CityFlow~\cite{tang2019cityflow} provides cross-camera
identities but no language, while CityFlow-NL~\cite{Feng2021CityFlowNL}
evaluates retrieval over pre-built trajectories rather than query-driven
tracking. CRTrack~\cite{chen2025cross}, the closest referring setting, uses
synchronized overlapping views of a single scene with appearance-based
queries. None therefore evaluates free-form, query-driven tracking across a
sequential, sparse-overlap camera network, where targets may disappear between
views and must retain a consistent identity.

We formalize this setting as \textbf{Referring Multi-Target Multi-Camera
Tracking (Refer-MTMCT)} and introduce \textbf{\ourdata}, built on CityFlow's
global cross-camera identity annotations. \ourdata augments these trajectories with
structured, spatio-temporally grounded language and supports many-to-many
query-to-trajectory mappings. It combines AI-generated queries for controlled coverage with real-user queries capturing natural ambiguity and an
evaluation protocol for language grounding, cross-camera association, and
spatio-temporal reasoning.

To establish end-to-end reference performance, we provide
\textbf{\ourmodel}, a modular multimodal reference framework in which an
LLM-based agent interprets and refines the query and orchestrates
interchangeable detection, tracking, and cross-camera association modules
(Fig.~\ref{fig:general_archi}). \ourmodel is a reference implementation for running and analyzing the benchmark; it is not a new tracking method.

\textbf{Contributions.}
\textbf{(i) Refer-MTMCT:} we formalize a new evaluation setting in which
free-form queries define targets to be tracked with consistent identities across
a camera network.
\textbf{(ii) \ourdata:} we introduce, to the best of our knowledge, the first benchmark for
this setting, with AI-generated and real-user language annotations, many-to-many
query-to-trajectory mappings, and a protocol evaluating language grounding,
cross-camera association, and spatio-temporal reasoning.
\textbf{(iii) \ourmodel:} we provide a modular reference framework that
establishes end-to-end reference performance and supports benchmark analysis
through detector, cross-camera association, interactive refinement, and
language-model studies.

\section{Related Work}
\label{sec:related_work}


\textbf{Object Tracking.} 
MOT maintains object identities within a single camera. Representative methods
include Tracktor~\cite{Tracktor}, CenterTrack~\cite{CenterTrack}, and later
approaches using transformer representations~\cite{TransCenter} or improved
association~\cite{aharon2022bot}. MTMCT extends tracking across cameras,
typically by combining single-camera tracking with cross-camera association
using appearance or spatio-temporal priors~\cite{yao2022city,ullah2018directed,tesfaye2017}.
Both settings assume that the targets of interest are defined independently of
a language query; Refer-MTMCT instead lets a free-form query define the target
set across cameras.



\textbf{Refer-MOT.} 
Refer-MOT replaces manual target selection with natural-language
queries~\cite{nguyen2024type,du2024ikun,ma2024mls}.
TransRMOT~\cite{wu2023referring} introduced the task on Refer-KITTI, and
TempRMOT~\cite{zhang2024bootstrapping} strengthens temporal modeling, but both
remain single-camera. CityFlow-NL~\cite{Feng2021CityFlowNL} attaches
trajectory-level captions to CityFlow~\cite{tang2019cityflow}, but evaluates
retrieval over pre-built trajectories and single-target localization rather
than query-conditioned multi-camera tracking. CRTrack~\cite{chen2025cross},
the closest referring setting, uses synchronized views of one scene
(Fig.~\ref{fig:intro}) rather than a sequential camera network.

\reviewedit{In Refer-MTMCT, the query itself defines which cross-camera trajectories form the target set, whereas retrieval-style systems rank trajectories built without the query.}
Such systems do not directly enter our HOTA/IDF1 protocol without an additional tracking and
cross-camera association stage; Sec.~\ref{sec:experiments} positions the
baselines accordingly.

\section{\ourdata Benchmark Dataset}
\label{sec:benchmark}

\begin{figure}[!tb]
\centering
\includegraphics[width=0.75\linewidth]{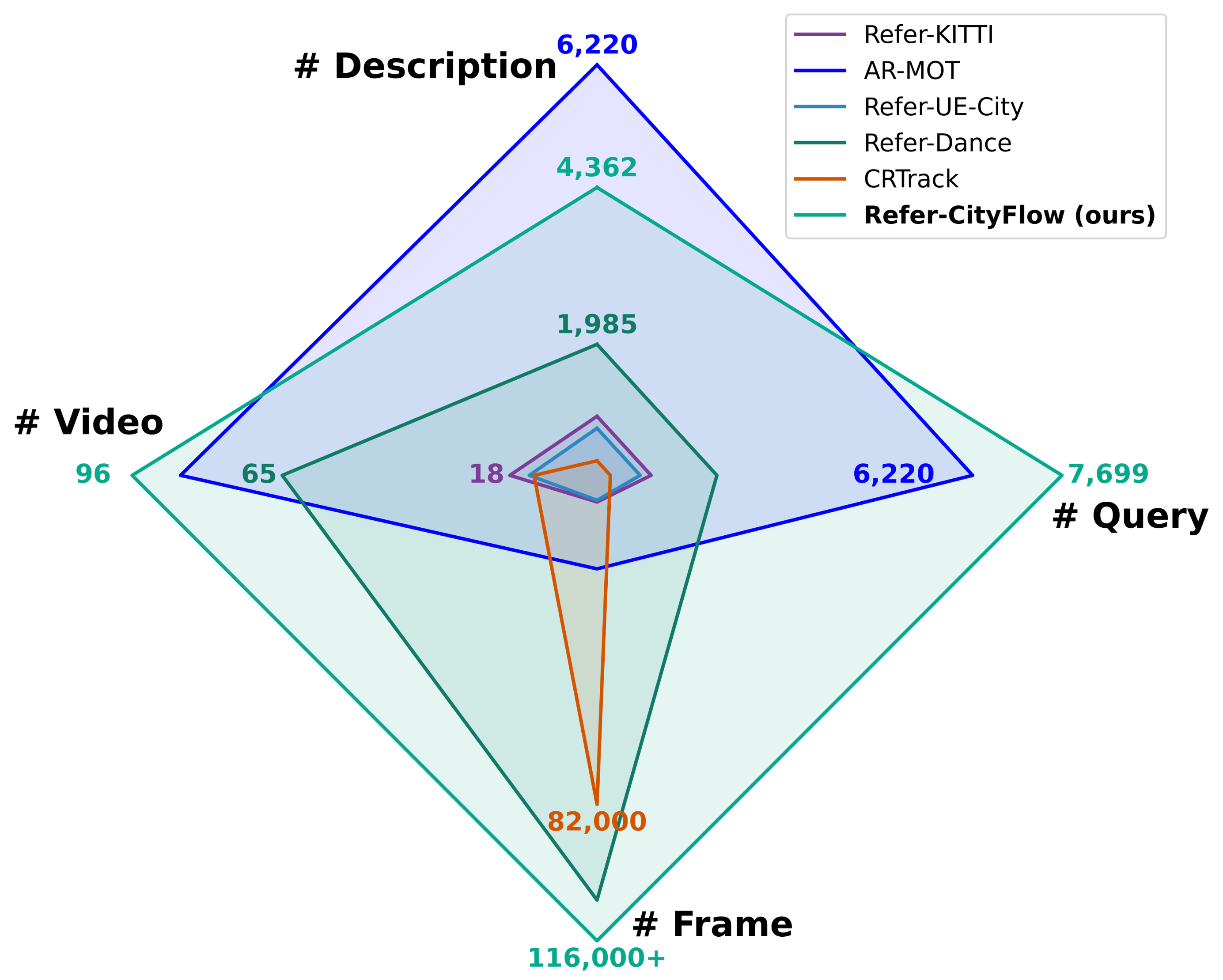}
\caption{\textbf{Quantitative dataset-scale comparison.}
Coverage of \ourdata and prior tracking datasets in frames, videos, queries,
and trajectory descriptions.}
\label{fig:radar_chart}
\vspace{-10pt}
\end{figure}

\ourdata is a benchmark dataset for Referring Multi-Target Multi-Camera Tracking
(Refer-MTMCT), built on CityFlow cross-camera vehicle trajectories and
augmented with structured semantic annotations and natural-language queries. Tab.~\ref{tab:dataset_stats} summarizes \ourdata and
Fig.~\ref{fig:radar_chart} places its scale relative to prior tracking datasets, while Tab.~\ref{tab:dataset_comparison} situates
\ourdata among prior referring datasets.
 
\begin{table}[t]
\centering
\setlength{\tabcolsep}{6pt}
\renewcommand{\arraystretch}{1.05}
\caption{\textbf{Summary statistics of \ourdata.}}
\label{tab:dataset_stats}
\scriptsize
\begin{tabular}{ccc}
\toprule
\textbf{Cameras / videos} & \textbf{Total frames} & \textbf{Global trajectories} \\
40 / 96 & 116{,}000+ & 660+ \\
\midrule
\textbf{Descriptions} & \textbf{Queries (AI / real-user)} & \textbf{Query-to-trajectory} \\
4{,}362 & 6{,}000 / 1{,}699 & many-to-many \\
\bottomrule
\end{tabular}
\end{table}


\subsection{Task and Benchmark Design}
\label{sec:task_def}

Given a free-form natural-language query $q$ and camera streams
$\{V_1,\ldots,V_N\}$, Refer-MTMCT requires a system to return a set of
identity-consistent trajectories $\{T_1,\ldots,T_k\}$, each spanning one or
more cameras, whose targets satisfy the semantic and spatio-temporal constraints
expressed in $q$. A query may correspond to one or more trajectories.

\textbf{Design principles.}
\ourdata is structured around four principles:
(i)~\emph{query-defined target sets}, rather than ranking pre-built tracks;
(ii)~\emph{cross-camera identity consistency} under viewpoint changes and
discontinuous observations;
(iii)~\emph{compositional spatio-temporal grounding} over appearance, motion,
location, and temporal constraints; and
(iv)~\emph{complementary language conditions}, combining AI-generated queries for
controlled coverage with real-user queries capturing natural ambiguity.

\textbf{Scope and assumptions.} Inheriting CityFlow restricts the object
domain to vehicles, so results on \ourdata concern referring tracking in traffic surveillance only. The benchmark does not prescribe a camera-overlap pattern or cross-camera association mechanism: predictions are scored against annotated trajectories, and identity may be recovered by any method. Overlap is required only by the reference associator in Sec.~\ref{sec:methodology}, not by the task or benchmark.

\begin{table}[t]
\caption{\textbf{Qualitative comparison of referring-tracking datasets.}
Prior referring-tracking datasets predominantly target single-camera or synchronized-view settings; \ourdata targets sequential multi-camera tracking with structured spatio-temporal and real-user queries. Refer-KITTI-V2 and the TempRMOT baseline share one source paper~\cite{zhang2024bootstrapping}. CityFlow-NL and
CRTrack are listed apart from Refer-MTMCT: the former ranks precomputed
trajectories rather than driving the tracking, and the latter observes one scene
from simultaneous overlapping views rather than a camera network.}
\label{tab:dataset_comparison}
\centering
\scriptsize
\newcommand{\tabrr}{\raggedright\let\\\tabularnewline}
\setlength{\tabcolsep}{2pt}
\renewcommand{\arraystretch}{1.15}
\begin{tabular}{@{} p{1.18cm} | p{1.38cm} p{1.16cm} p{1.02cm} p{1.32cm} p{1.12cm} @{}}
\toprule
 \tabrr \textbf{Type} &  \tabrr \textbf{Dataset} &  \tabrr \textbf{Scenes} &  \tabrr \textbf{Cameras} &  \tabrr \textbf{Queries} &  \tabrr \textbf{Language} \\
\midrule
\multirow{4}{*}{Refer-MOT}
  &  \tabrr Refer-KITTI \citep{wu2023referring} &  \tabrr Street-level vehicles only &  \tabrr On-vehicle &  \tabrr Spatial &  \tabrr 49 unique words \\
\cmidrule{2-6}
  &  \tabrr Refer-KITTI-V2 \citep{zhang2024bootstrapping} &  \tabrr Street-level driving &  \tabrr On-vehicle &  \tabrr Spatial, motion &  \tabrr Wider than Refer-KITTI \\
\cmidrule{2-6}
  &  \tabrr Refer-UE-City \citep{ma2024mls} &  \tabrr Synthetic urban &  \tabrr Roadside &  \tabrr Spatial, action &  \tabrr Limited style \\
\cmidrule{2-6}
  &  \tabrr Refer-Dance \citep{du2024ikun} &  \tabrr Indoor dancing &  \tabrr Dancing &  \tabrr Group motion &  \tabrr 48 distinct expr. \\
%
%
\midrule
 \tabrr Multi-cam retrieval &  \tabrr CityFlow-NL \citep{Feng2021CityFlowNL} &  \tabrr Urban traffic &  \tabrr Roadside &  \tabrr Appearance, motion (retrieval) &  \tabrr Trajectory captions \\
\midrule
 \tabrr Cross-view referring &  \tabrr CRTrack \citep{chen2025cross} &  \tabrr Outdoor pedestrian &  \tabrr Multi-view, one scene &  \tabrr Appearance only &  \tabrr 221 descriptions \\
\midrule
 \tabrr Refer-MTMCT &  \tabrr \textbf{\ourdata{} (ours)} &  \tabrr \textbf{Urban traffic} &  \tabrr \textbf{Roadside network} &  \tabrr \textbf{Appearance, direction, zone, action, relation, lane, spatial, temporal} &  \tabrr \textbf{6,000 AI-generated, 1,699 human requests} \\
\bottomrule
\end{tabular}
\end{table}

\subsection{Source Data and Annotation}
\label{sec:data_source}
\label{sec:data_structure}

\ourdata builds on CityFlow~\cite{tang2019cityflow}, a public city-scale
multi-camera traffic benchmark providing synchronized roadside video, global
vehicle identities across non-overlapping and partially overlapping views, and
per-camera homographies. These existing cross-camera identities allow our added
language annotations to be linked directly to global trajectories rather than
re-deriving identity ground truth. \ourdata retains CityFlow's imagery and
global IDs while adding semantic attributes, zone definitions,
trajectory-level descriptions, and natural-language queries.

Each trajectory is annotated along four attribute groups:
(1)~\emph{appearance} (color and vehicle category),
(2)~\emph{motion} (direction and time-aligned actions),
(3)~\emph{spatial context} (camera, zone, and lane), and
(4)~\emph{relations} to nearby objects.
Queries use a structured schema covering color, vehicle type, camera ID,
direction, zone, action, lane, relation, time range, and optional reappearance
cameras. A query constrains a subset of these fields, producing the six
AI-generated formats described in Sec.~\ref{sec:data_ai_query}. Most fields use
controlled value sets, while vehicle type may also contain free-text values.
The complete schema and per-camera zoning definitions are provided in the
supplementary material.

\textbf{Annotation protocol and ground truth.}
All 4{,}362 trajectory-level descriptions are manually annotated under a
common guideline covering vehicle attributes, zones, lanes, directions, and
temporal information. Four annotators review each target's multi-camera clip
and revise its description accordingly, requiring 960 annotation hours in
total. The annotations are then cross-checked for inconsistent, ambiguous, or
missing information, followed by a 120-hour final quality-control pass by an
additional reviewer to verify annotation consistency and query-to-trajectory
correspondence.

Each query is associated with one or more target trajectories through its semantic and spatio-temporal constraints, without manual enumeration.
For AI-generated queries, these constraints are represented in structured form
and matched against the annotated trajectory database
(Sec.~\ref{sec:data_ai_query}), linking the added semantic annotations to
CityFlow's global cross-camera identities. Queries with insufficient matching
support are filtered during generation, while ambiguous or multi-target cases
are additionally inspected manually before inclusion.

\subsection{Natural-Language Query Construction}
\label{sec:query_construction}
\label{sec:data_queries}

\ourdata provides two complementary query sets:
(1)~\emph{AI-generated queries} for scalable and systematic benchmarking,
and (2)~\emph{real-user queries} capturing natural linguistic variation,
ambiguity, and open-vocabulary expressions.
We distinguish \emph{trajectory-level descriptions} from \emph{queries}: descriptions
annotate individual trajectories, whereas queries are system inputs and may
match one or multiple trajectories. This creates a many-to-many
query-to-trajectory mapping, unlike the one-to-one trajectory-caption setting
used by CityFlow-NL~\cite{Feng2021CityFlowNL}.


\subsubsection{AI-Generated Query Set}
\label{sec:data_ai_query}

We define six query formats, each centered on one capability from the attribute
schema (Tab.~\ref{tab:query_formats}). The formats are compositional: a query
may combine its primary capability with cues from other formats and additional
spatial or temporal constraints. We generate 1{,}000 queries per format,
yielding 6{,}000 queries in total.


\begin{table}[t]
\centering
\setlength{\tabcolsep}{0pt}
\renewcommand{\arraystretch}{1.15}
\caption{\textbf{The six AI-generated query formats.} Each format is built
around one constraint type; queries may combine several.}
\label{tab:query_formats}
\scriptsize
\newcommand{\ftabrr}{\raggedright\let\\\tabularnewline}
\begin{tabular}{@{}p{0.95cm}@{\hspace{6pt}}p{1.85cm}@{\hspace{6pt}}p{4.60cm}@{}}
\toprule
\textbf{Format} & \textbf{Capability} & \textbf{Example constraints} \\
\midrule
QF-1 & \ftabrr Appearance & \ftabrr Color, vehicle type \\
QF-2 & \ftabrr Direction  & \ftabrr Motion direction \\
QF-3 & \ftabrr Zone       & \ftabrr Entry, exit, or transition zone \\
QF-4 & \ftabrr Action     & \ftabrr Going straight, turning, stopping \\
QF-5 & \ftabrr Relation   & \ftabrr Relation to a nearby object \\
QF-6 & \ftabrr Lane       & \ftabrr Lane-level position \\
\bottomrule
\end{tabular}
\end{table}

\textbf{Generation process.} 
Queries are generated from the annotated ground-truth database so that their
target sets are known. Per-vehicle attributes, including color, type, camera,
direction, zone, action, relation, lane, and time range, are provided to an LLM
that generates temporal query variants, each requiring the target to reappear
in at least one additional camera. Every generated query is then executed
against the database for validation, and queries with insufficient support are
discarded. Examples from both query sets appear in
Tab.~\ref{tab:query_examples}; additional per-format examples and generation
prompts are provided in the supplement.

\begin{table}[!tb]
\caption{\textbf{Example AI-generated and real-user queries.}
The examples illustrate different query constraints, including an
underspecified real-user query and its clarified form.}
\label{tab:query_examples}
\centering\scriptsize
\renewcommand{\arraystretch}{1.15}
\newcommand{\qtabrr}{\raggedright\let\\\tabularnewline}
\setlength{\tabcolsep}{0pt}
\begin{tabular}{@{}p{1.42cm}@{\hspace{6pt}}p{1.72cm}@{\hspace{6pt}}p{4.60cm}@{}}
\toprule
\textbf{Source} & \textbf{Type} & \textbf{Query} \\
\midrule
\qtabrr AI-generated
  & \qtabrr Appearance + reappearance
  & \qtabrr ``Track all white trucks that appear in camera~16, and also appear at
    camera~17.'' \\
\addlinespace[4pt]
  & \qtabrr Zonal + temporal (compositional)
  & \qtabrr ``Track all the black sedans that are moving from zone~22d to zone~22b in
    camera~22 from 15.8 seconds to 25.3 seconds, and also appear at
    cameras~16,17,18.'' \\
\midrule
\qtabrr Real-user
  & \qtabrr Raw, underspecified
  & \qtabrr ``Track all green minivans that move from camera~6 to camera~9 now
    and yesterday from 2~PM to 5~PM.'' \\
\addlinespace[4pt]
  & \qtabrr Clarification and resolved form
  & \qtabrr System: ``Do you mean the cameras in the center of Chattanooga?'' User:
    ``Yes, Chattanooga Center.'' Resolved fields: minivan, green; path 6, 7, 8, 9;
    current and yesterday 2--5~PM. \\
\bottomrule
\end{tabular}
\end{table}

\subsubsection{Real-User Query Set}
\label{sec:data_real_query}

\begin{figure}[h]
\centering
\includegraphics[width=\columnwidth]{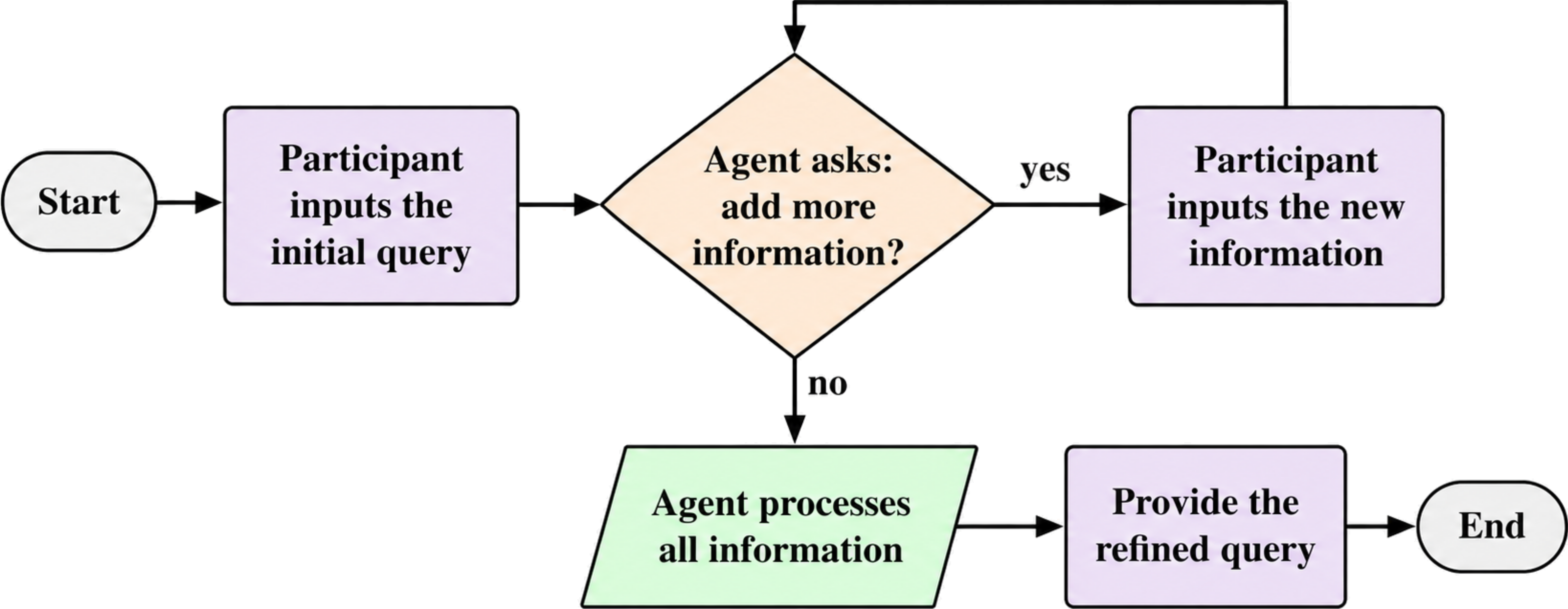}
\caption{\textbf{Real-user query collection and refinement.} Participants freely
describe a target; when the query is underspecified, the agent asks targeted
follow-up questions until it can be resolved into an executable form, which is
run against the ground-truth database to obtain the target set. Both the raw and
resolved queries are retained for evaluation.}
\label{fig:human_query}
\end{figure}

We collect 1{,}699 real-user queries from 50 participants with diverse
backgrounds (students, computer scientists, transportation professionals,
managers). Participants freely describe vehicles they wish to track, giving
queries with more variable language and intent (incomplete specifications,
ambiguous wording, open-vocabulary expressions) than the AI-generated set.
Because such queries are often underspecified, the benchmark pairs raw queries
with refined, executable forms obtained through interactive clarification
(Fig.~\ref{fig:human_query}): the system asks targeted follow-up questions to fill
in missing attributes, and the resolved query is executed against the
ground-truth database to obtain the target set. Both the raw user query and its
refined executable form are retained; the matching trajectories returned by the
refined form constitute the evaluation ground truth. Representative before/after
refinement examples are provided in the supplement.

\subsection{Benchmark Evaluation Protocol}
\label{sec:experiments_protocol}

The protocol scores the trajectories returned for each query against its
annotated target set. Because target membership is query-defined, a query may
have multiple positives and a trajectory may satisfy multiple queries. The
protocol preserves this many-to-many relation and is model-agnostic: any system
producing query-conditioned cross-camera trajectories can be evaluated.


\textbf{Metrics.}
We report HOTA, DetA, and AssA~\cite{luiten2021hota} together with
IDF1~\cite{ristani2016performance}, covering detection quality, association,
and identity consistency. The six AI-generated query formats provide
capability-specific breakdowns, while the 30-query human comparison
uses mean L1 target-count error. Formal metric definitions are provided in the
supplement.


\textbf{Scoring procedure.}
Predictions are evaluated per (query, camera) pair and matched to ground truth by IoU, using the standard HOTA threshold range and $0.5$ for IDF1. Metrics are micro-averaged by pooling true positives, false positives, and false negatives across all pairs before scoring, so longer or multi-target queries contribute proportionally. Queries with no returned predictions remain in the evaluation and contribute false negatives rather than being skipped. Because identity matching is performed within each pair, these scores do not penalize a correctly returned vehicle that receives different labels in different cameras; the supplement reports a global-identity IDF1 for this case (Tab.~S3) and the exact submission format.

\section{\ourmodel: A Modular Reference Framework}
\label{sec:methodology}

\begin{figure*}[t]
\begin{center}
	\includegraphics[width=\linewidth]{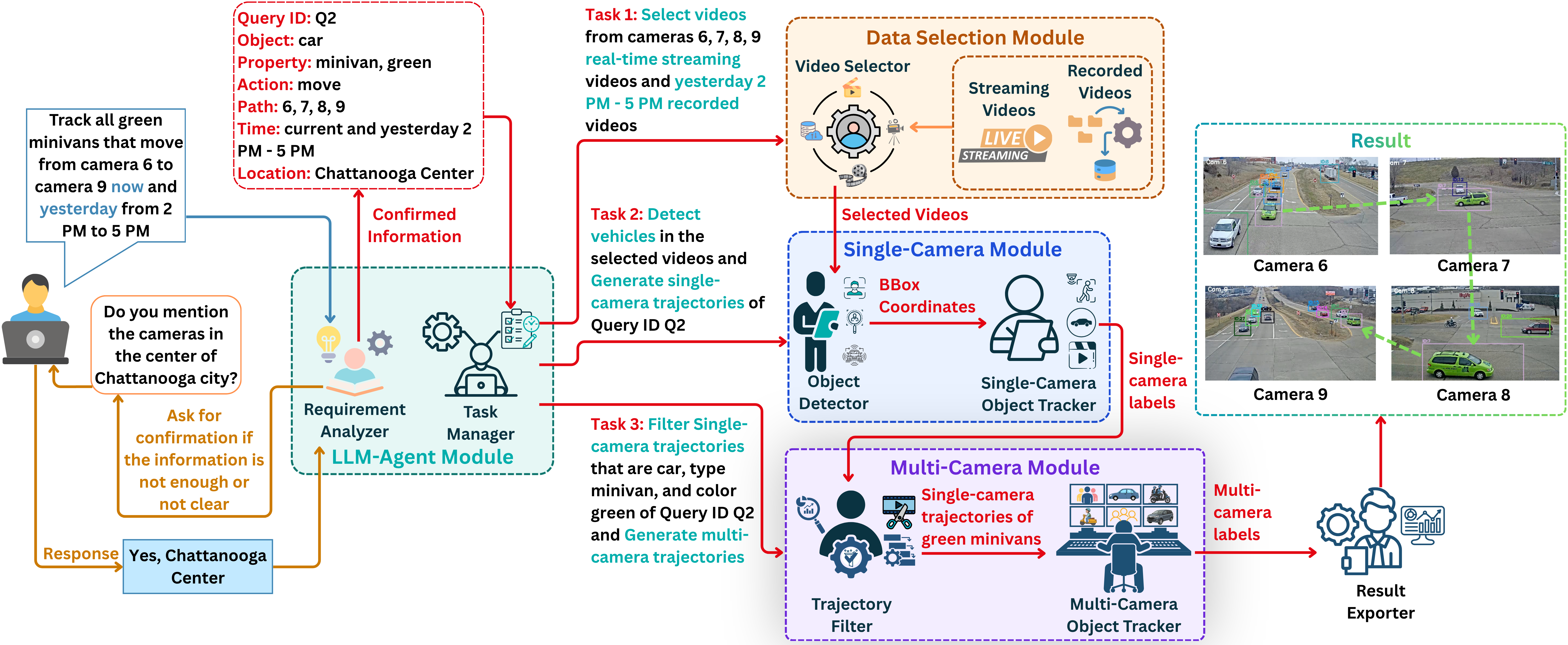}
        \caption{\textbf{A sample query and overview of the \ourmodel framework.}
The LLM-Agent Module parses the query and, when it is underspecified, interactively clarifies it with the user (\textcolor{myblue}{blue}: user query; \textcolor{myorange}{orange}: the analyzer requests missing information; \textcolor{myred}{red}: agent interaction). The resolved query then drives video and time-window selection, single-camera perception, and cross-camera trajectory association.
        \vspace{-10pt}}
    \label{fig:general_archi}
\end{center}
\end{figure*}

\ourmodel is a modular multimodal framework connecting language reasoning with
multi-camera perception. We present it as a reference framework for running
\ourdata end-to-end; each component is replaceable without changing the evaluation protocol. An LLM-based coordination module orchestrates four stages
(Fig.~\ref{fig:general_archi}): (i) query interpretation and task planning,
(ii) video selection, (iii) single-camera perception, and (iv) cross-camera
trajectory association. Inspired by MetaGPT~\cite{metagpt}, this design separates language reasoning from visual
execution while allowing individual perception and association components to
be exchanged independently.


\subsection{LLM-Agent Module}
The LLM-Agent Module converts a free-form request into a structured,
executable query through two collaborative agents: the
\textbf{Requirement Analyzer} and the \textbf{Task Manager}. Their operations
are text-based; visual input is handled by downstream perception modules, so
any sufficiently capable LLM can serve this role
(Tab.~\ref{tab:llm}).


\textbf{Requirement Analyzer and interactive refinement.}
The Requirement Analyzer parses the request into a structured
\textit{Requirement Schema} containing target categories, fine-grained
attributes, spatial constraints, and temporal bounds. For ambiguous or
underspecified queries, it identifies missing constraints and asks targeted
natural-language questions before execution.

\textbf{Task Manager.}
The Task Manager organizes pipeline execution from the Requirement Schema
through model selection, task decomposition, and adaptive re-execution.
Standard categories are routed to efficient detectors such as
YOLO26~\cite{yolo26_ultralytics}, whereas complex or long-tail descriptions
can use vision--language models such as
Grounding-DINO~\cite{ren2024grounding} or
OmDet-Turbo~\cite{zhao2024real}. Multi-camera requests are decomposed into
camera- and time-specific subtasks for parallel execution. When intermediate
results fail to satisfy the query constraints, the manager can adjust
processing parameters or expand the search space before returning a result.

\subsection{Data Selection Module}

The module identifies the video sources relevant to the
structured query produced by the LLM-Agent Module. As illustrated in
Fig.~\ref{fig:general_archi}, it includes a \textit{Video Selector}
that determines which video streams should be processed for tracking. Given
query constraints such as camera IDs, time intervals, and spatial conditions,
the Video Selector dynamically selects between \textit{recorded videos} and
\textit{streaming videos}. For queries involving historical time ranges, the
module retrieves the corresponding segments from recorded video data. For
real-time queries, it selects active streaming feeds and continuously filters
incoming frames. The selected videos are then forwarded to the Single-Camera
Module for object detection and trajectory extraction.

\subsection{Single-Camera Module: Perception and Attribute Filtering}
\label{sec:single_camera}
Given the selected video streams, this module produces per-camera tracklets with
associated appearance and attribute information. Following the Task Manager's
dispatch, it selects one of five detectors according to the Requirement Schema
and performs tracking with OC-SORT.

\textbf{Multi-model object detection.}
\ourmodel supports two complementary detector families.
Efficiency-oriented YOLO models~\cite{yolo26_ultralytics} (YOLOv5 and YOLO26)
handle common object categories with high inference speed, while
open-vocabulary models such as Grounding-DINO~\cite{ren2024grounding},
OmDet-Turbo~\cite{zhao2024real}, and Qwen3-VL~\cite{bai2025qwen3} handle
rarer or more complex language descriptions without task-specific training.
The Task Manager selects between these alternatives according to the parsed
query requirements.

\textbf{Localization and temporal association.}
Given the detections, OC-SORT~\cite{cao2023observation}, augmented with
camera-motion compensation (CMC), associates bounding boxes over time to
maintain tracklet continuity under occlusion and motion.

\textbf{Linguistic attribute filtering.}
For closed-vocabulary detectors, object crops are embedded using a feature
extractor~\cite{zhou2021mixstyle} and classified by linear SVMs for vehicle
color and type~\cite{szucs2023hierarchical}; tracklets inconsistent with the
query are then removed. Open-vocabulary detectors instead encode attribute
constraints directly in their prompts and bypass this filtering stage. This
supports both classifier-based and prompt-based grounding within the same
reference framework.





\subsection{Multi-Camera Module: Global Trajectory Synthesis}
\label{sec:multi_camera_module}

The Multi-Camera Module merges per-camera tracklets into globally consistent
trajectories. This associator is one possible choice and is not part of the task definition: alternative associators, including appearance-based
re-identification methods that do not require camera overlap, can be evaluated
under the same protocol.
Our implementation builds on the fast online MTMCT framework of
Shim et al.~\cite{shim2023fast} and adapts it to Refer-MTMCT through
(1) automatic geometric and temporal calibration,
(2) query-aware primary-camera anchoring, and
(3) constraint-driven trajectory filtering.


\smallskip

\textbf{Geometric and temporal pre-processing.}
The base MTMCT method requires camera-overlap masks and temporal alignment.
Rather than manually specifying them, we derive overlap regions from the
dataset-provided homographies $\mathbf{H}_i$ and estimate a canonical timeline
from GPS-matched co-observed vehicles. Depending on the number of matched
samples, temporal alignment uses isotonic regression, a median offset, or
timestamp interpolation; full construction details are provided in the
supplement. On CityFlow Scenario~02, the automatically generated masks reach
IoU $0.80$--$0.97$ against manual annotations, with less than $0.1\%$
difference in the resulting IDF1.


\smallskip

\textbf{Cross-camera association.}
At each canonical time, active tracklets are represented by their
highest-confidence appearance features. Pairwise cosine distances are computed
after excluding same-camera and overlap-invalid pairs, and the remaining
tracklets are grouped by complete-linkage clustering with the threshold selected
by the Dunn index~\cite{dunn1974}. Existing global IDs are propagated within
clusters; remaining tracklets are matched to lost trajectories using the
Hungarian algorithm~\cite{munkres1957}, and unmatched tracklets receive new
global IDs. Assigned global IDs remain fixed to preserve temporal consistency.


\smallskip


\textbf{Query-Anchored Trajectory Filtering.} We introduce a
\textbf{primary-camera anchor mechanism}: a global trajectory is kept only if at
least one of its tracklets appears in the query-specified primary camera within
the given time window, while auxiliary cameras stay unconstrained so the system
can still recover the full trajectory. Unlike conventional MTMCT that operates independently of user intent,
this integrates query constraints into global trajectory selection,
restricting the returned trajectories to those consistent with the
query-specified primary-camera and temporal constraints.

\section{Experiments}
\label{sec:experiments}


\subsection{Experimental Setup}
\textbf{Datasets \& Metrics.} We evaluate on \ourdata (Sec.~\ref{sec:benchmark})
and, for the conventional setting, on CityFlow~\cite{tang2019cityflow}. All
tracking results in Tabs.~\ref{tab:res_ai_query} and~\ref{tab:res_real_query} use
Scenario~02 (S02) of both datasets, a four-camera scenario also used in prior
MTMCT evaluations~\cite{herzog2024spatial, huang2025online, shim2023fast}. The complete \ourdata benchmark spans 40
cameras, so these numbers evaluate the reference methods on S02 rather than
constituting a full-network evaluation. We follow the evaluation protocol in Sec.~\ref{sec:experiments_protocol}.

\textbf{Baselines.} Our baselines are TransRMOT~\cite{wu2023referring} and
TempRMOT~\cite{zhang2024bootstrapping}, with their publicly released weights.
Both are designed for single-camera settings and cannot perform cross-camera
association, so they primarily indicate performance in single-camera referring
tracking; they are scored under the same protocol as \ourmodel
(Sec.~\ref{sec:experiments_protocol}). Two-stage plug-in referrers such as
iKUN~\cite{du2024ikun} match text against per-camera tracklets and do not emit
cross-camera identities; adapting them to Refer-MTMCT would require a new
association stage, which we leave to future work.

\textbf{Implementation Details.} The LLM-Agent module is implemented
with OpenAI gpt-4o-mini \reviewedit{with temperature set to~0;}
the same model serves as the
Requirement Analyzer and Task Manager and produces the results in
Tabs.~\ref{tab:res_ai_query} and ~\ref{tab:res_real_query}. The
provider comparison in Tab.~\ref{tab:llm} varies only this parsing model
and leaves the rest of the pipeline fixed. For
closed-vocabulary detection, we use YOLO26-m and YOLOv5 with linear SVM
classifiers for vehicle color and type; for open-vocabulary grounding, we use
Grounding-DINO-1.5~\cite{ren2024grounding}, OmDet-Turbo~\cite{zhao2024real}, and
Qwen3-VL-8B~\cite{bai2025qwen3} (confidence threshold 0.25).
\textit{Single-camera tracking} uses OC-SORT~\cite{cao2023observation} with a
detection threshold of 0.2, maximum age of 30 frames, and minimum hits of 3.
\textit{Query-conditioned processing.} The time window of a query is applied
strictly to the primary camera and extended by $\pm10$\,s on auxiliary cameras
to capture cross-camera transitions. \textit{Cross-camera association} uses a
cosine distance threshold of $\tau_{\text{mtmc}}=0.65$ and a maximum inter-frame
gap of 30 frames. All experiments run on a single NVIDIA~A100-80GB GPU. No component is trained or fine-tuned on \ourdata. The detectors, the tracker,
and the parsing model are used as publicly released, and the re-identification
feature extractor is taken from the released Fast Online MTMCT
implementation~\cite{shim2023fast}. All results reported here are therefore zero-shot with respect to the benchmark.

\subsection{Experimental Results}

\begin{table}[t]
\centering
\caption{\textbf{Performance on AI-generated queries in \ourdata Scenario~02.} ``SVM'': classifier for attribute filtering (att.\ fil.).}
\label{tab:res_ai_query}
\scriptsize
\setlength{\tabcolsep}{1.0pt}
{\resizebox{\columnwidth}{!}{\begin{tabular}{l|l|c|rrrr}
\toprule
\multicolumn{1}{c|}{\textbf{Model}} &
\multicolumn{1}{c|}{\textbf{Detector}} & \textbf{Training}
& 
\multicolumn{1}{c}{\textbf{HOTA}} &
\multicolumn{1}{c}{\textbf{DetA}} &
\multicolumn{1}{c}{\textbf{AssA}} &
\multicolumn{1}{c}{\textbf{$\text{IDF1}$ }} \\ \toprule
TransRMOT \cite{wu2023referring} & Def.\ DETR & MS COCO & 8.55 & 3.19  & 24.75 & 3.54 \\ 
TempRMOT \cite{zhang2024bootstrapping} & Def.\ DETR & MS COCO & 9.19 & 3.27 & 31.94 & 3.16 \\ \midrule
\multirow{5}{*}{\textbf{\ourmodel (ours)}} & YOLOv5+SVM  & COCO\,+\,att.\ fil. & 24.15 & 13.39 & 48.65 & 26.28 \\
& YOLO26-m+SVM     & COCO\,+\,att.\ fil. & 26.79 & 17.37 & 45.82 & 32.13 \\ \cmidrule{2-7}
 & Qwen3-VL   & Open-vocab.  & 29.03 & 19.30 & 47.84 & 38.58 \\ 
& Grounding-DINO  & Open-vocab. & 28.76 & 17.42 & 49.38 & 37.72 \\ 
 & OmDet-Turbo     & Open-vocab. & \textbf{31.08} & \textbf{19.58} & \textbf{51.08} & \textbf{43.18} \\ \bottomrule
\end{tabular}}}
\end{table}


\noindent\textbf{How far do existing methods get?} On the six AI-generated query formats (Tab.~\ref{tab:res_ai_query}), TransRMOT and TempRMOT reach only 8.55 and 9.19 HOTA, with IDF1 below 4.0. As single-camera methods without cross-camera association, their scores are a single-camera reference point, not a matched comparison.
\reviewedit{\ourmodel with OmDet-Turbo achieves 31.08 HOTA and 43.18 IDF1 without retraining.}

\begin{table}[t]
\centering
\caption{\textbf{Performance on real-user queries (open-vocabulary, human-authored) in \ourdata Scenario~02.} Single-camera baselines and the cross-view CRTrack are run directly on \ourdata. ``SVM'': classifier for attribute filtering (att.\ fil.). See the supplement for a positional-subset ablation.}
\label{tab:res_real_query}
\scriptsize
\setlength{\tabcolsep}{1.0pt}
\resizebox{\columnwidth}{!}{\begin{tabular}{l|l|c|rrrr}
\toprule
\multicolumn{1}{c|}{\textbf{Model}} &
\multicolumn{1}{c|}{\textbf{Detector}} & \textbf{Training}
& 
\multicolumn{1}{c}{\textbf{HOTA}} &
\multicolumn{1}{c}{\textbf{DetA}} &
\multicolumn{1}{c}{\textbf{AssA}} &
\multicolumn{1}{c}{\textbf{$\text{IDF1}$ }} \\ \toprule
TransRMOT \cite{wu2023referring} & Def.\ DETR & MS COCO & 18.28 & 10.38 & 33.96 & 7.62 \\ 
TempRMOT \cite{zhang2024bootstrapping} & Def.\ DETR & MS COCO & 14.95 & 6.46 & 37.49 & 9.27 \\
CRTrack~\cite{chen2025cross} & APTM & Frozen & $\sim$0.05 & $\sim$0.00 & $\sim$1.55 & $\sim$0.00 \\ \midrule
\multirow{5}{*}{\textbf{\ourmodel (ours)}} & YOLOv5+SVM  & COCO\,+\,att.\ fil. & 42.93 & 34.15 & 56.19 & 56.15 \\
& YOLO26-m+SVM     & COCO\,+\,att.\ fil. & 34.56 & 27.58 & 47.67 & 43.98 \\ \cmidrule{2-7}
 & Qwen3-VL   & Open-vocab.  & 43.30 & 36.20 & 53.95 & 64.75 \\ 
  & Grounding-DINO     & Open-vocab. & \textbf{44.10} & \textbf{37.35} & 54.35 & \textbf{65.40} \\
& OmDet-Turbo  & Open-vocab. & 40.70 & 29.95 & \textbf{57.30} & 58.15 \\ 
 \bottomrule
\end{tabular}}
\end{table}


\noindent Results on real-user queries (Tab.~\ref{tab:res_real_query}) show the same trend: single-camera baselines reach up to 18.28 HOTA, whereas \ourmodel variants achieve at least 34.56 HOTA. Grounding-DINO performs best overall with 44.10 HOTA and 65.40 IDF1, while YOLOv5+SVM remains competitive despite using a closed vocabulary. 
\reviewedit{Scores are higher on real-user queries, although differences in query composition limit direct comparison between the two sets.}
\noindent
\reviewedit{CRTrack~\cite{chen2025cross} achieves $\sim$0.05 HOTA when transferred directly to \ourdata. CRTrack was built for pedestrians in synchronized overlapping views and does not transfer to vehicles moving through a sequential camera network.}

\textbf{Component sensitivity.} We vary selected components while
holding the rest of the pipeline fixed to assess their effect on performance.
Enabling the Multi-Camera Module raises IDF1 by $+$7.3
(YOLOv5+SVM) and $+$10.9 (YOLO26-m+SVM) and reduces identity switches by
approximately half, while per-camera IDF1 remains stable, showing that
cross-camera association improves identity consistency.


Varying the detector alone, with all other components fixed and without retraining, yields 24.15--31.08 HOTA on AI-generated queries and 34.56--44.10 HOTA on real-user queries. \reviewedit{The choice of detector therefore has a large effect.}
Varying the parsing LLM across four open and proprietary models
shifts field-exact accuracy by less than three points (97.1--100\%,
Tab.~\ref{tab:llm}) at a mean latency of 1.6\,s per query. Parsing errors are therefore rare on the evaluated QF-2 queries. We fix the model at
temperature~0, and fall back to a rule-based parser when the model output is malformed. The single-camera tracker and
cross-camera associator are held fixed to isolate the effects of the detector and parsing LLM.

\begin{table}[t]
\centering
\caption{\textbf{LLM parsing accuracy and estimated cost} across four open and proprietary models, measured on the directional (QF-2) query set; a query counts as correct only if every extracted field matches the ground truth. }
\label{tab:llm}
\scriptsize
\setlength{\tabcolsep}{6pt}
\renewcommand{\arraystretch}{0.95}
\begin{tabular}{l|rr}
\toprule
\textbf{Model} & \textbf{Accuracy} & \textbf{Est.\ Cost (USD)} \\
\midrule
Amazon Nova Micro & 99.4\%  & 0.110 \\
DeepSeek V3.2     & 97.1\%  & 1.828 \\
Meta Llama 3.1 8B & 100.0\% & 0.609 \\
GPT 4o-mini       & 100.0\% & 1.690 \\
\bottomrule
\end{tabular}
\end{table}


\textbf{Sanity checks.}
Two additional controls place the main
results in context. Under conventional MTMCT on CityFlow~\cite{tang2019cityflow},
\ourmodel with YOLOv5 reaches 86.84 IDF1 compared with 81.95 for Shim et
al.~\cite{shim2023fast}, providing a conventional-MTMCT reference point for the
same implementation. On a 30-query real-user subset, \ourmodel with
Grounding-DINO obtains a counting error comparable to the human reference
(30.15 vs.\ 30.50 L1) while requiring less processing time (948\,s vs.\
5{,}268\,s). Full results are provided in the supplementary material.

\subsection{Ablation Study}

The interactive refinement ablation (full table in the supplementary material)
evaluates interactive query refinement on the subset of real-user queries
that carry spatial (positional) constraints (not directly comparable to
Tab.~\ref{tab:res_real_query}) with the Grounding-DINO backbone. Removing it lowers
HOTA from 47.4 to 35.6, AssA from 59.6 to 41.3, and IDF1 from 69.3 to 51.2. 

\textbf{Error sensitivity.} 
\reviewedit{Parsing is field-exact on 97.1--100\% of QF-2 queries (Tab.~\ref{tab:llm}). On this subset, removing interactive refinement changes AssA and IDF1 more than DetA, so its main effect is on association.}
Detection shows the largest spread among the components we varied, and
a target missed per-camera cannot be recovered later. Ablating the Multi-Camera
Module moves IDF1 and identity switches while per-camera IDF1 is unchanged.
Because these interventions differ in scope and evaluation subset, their score
changes should not be read as shares of total end-to-end error.

\subsection{Qualitative Analysis}
The supplementary material traces the agent pipeline on a real
underspecified query, ``Find black vehicle go straight''. The Requirement
Analyzer extracts the stated fields, asks four targeted clarification
questions, and resolves the request into the paper's running example, a black
SUV from camera~6 reappearing in cameras~7 and~9; the downstream stages then
track the target across the three cameras over the resolved 5--60\,s window.

\section{Discussion}
\label{sec:discussion}

\textbf{Limitations.} 
\ourdata inherits its imagery, camera calibration, vehicle domain, and any
residual cross-camera identity errors from CityFlow. It therefore evaluates
English-language referring requests for vehicles within a single urban-traffic
corpus and does not establish generalization to pedestrians, indoor scenes,
multiple cities, multilingual requests, or substantially different camera
networks. In addition, the AI-generated query set is generated from a structured
attribute database and may exhibit linguistic regularities that are less common
in unconstrained human requests. We partly mitigate this limitation with the
complementary real-user query set, but broader domains, languages, and camera
topologies remain important directions for future extensions. Our reference experiments are restricted to CityFlow Scenario~02 rather
than the full 40-camera network; full-network evaluation remains an
important direction for future benchmark studies.

\textbf{Broader Impact and Ethics.} Language-driven multi-camera tracking can support situational awareness but also raises risks related to privacy, unauthorized surveillance, and biases in language and visual perception. \ourdata is built on public CityFlow data and focuses on vehicles rather than people, although bystanders may appear in the original footage. We release the benchmark for research use with documented licensing and intended-use guidelines, and recommend human oversight and appropriate access controls for deployment. Additional participant, licensing, and dataset documentation is provided in the supplementary material.

\section{Conclusion}

We introduced \textbf{Refer-MTMCT}, a new setting for language-driven multi-target multi-camera tracking, and \textbf{Refer-CityFlow}, a benchmark pairing free-form queries with identity-consistent trajectories across a camera network. We further developed M3Track, a modular multimodal reference framework
that combines LLM-based language reasoning with detection, tracking, and
cross-camera association. Reference experiments on Scenario~02 show that existing referring trackers transfer poorly to this setting, while M3Track establishes end-to-end reference performance.
Future work can advance this setting in two directions. First, Refer-CityFlow can be extended beyond urban vehicles and English queries to additional object categories, environments, and languages, enabling broader evaluation of Refer-MTMCT. Second, M3Track can incorporate stronger tracking and cross-camera association methods, particularly approaches that handle sparse or non-overlapping camera networks without relying on overlapping fields of view.

\section*{Acknowledgment}
This work was supported in part by the Transportation Network Growth Opportunity (TNGO) initiative funded by the Tennessee Department of Economic and Community Development, in collaboration with the University of Tennessee at Chattanooga and industry partners.


{\small
\bibliographystyle{ieeenat_fullname}
\bibliography{M3Track-ref}

\begin{thebibliography}{34}
\providecommand{\natexlab}[1]{#1}
\providecommand{\url}[1]{\texttt{#1}}
\expandafter\ifx\csname urlstyle\endcsname\relax
  \providecommand{\doi}[1]{doi: #1}\else
  \providecommand{\doi}{doi: \begingroup \urlstyle{rm}\Url}\fi

\bibitem[Aharon et~al.(2022)Aharon, Orfaig, and Bobrovsky]{aharon2022bot}
Nir Aharon, Roy Orfaig, and Ben-Zion Bobrovsky.
\newblock Bot-sort: Robust associations multi-pedestrian tracking.
\newblock \emph{arXiv preprint arXiv:2206.14651}, 2022.

\bibitem[Anh et~al.(2024)Anh, Tran, and Le]{anh2024tp}
Duy Le~Dinh Anh, Kim~Hoang Tran, and Ngan~Hoang Le.
\newblock Tp-gmot: Tracking generic multiple object by textual prompt with
  motion-appearance cost (mac) sort.
\newblock \emph{European Conference on Artificial Intelligence ECAI}, 2024.

\bibitem[Bai et~al.(2025)Bai, Cai, Chen, Chen, Chen, Cheng, Deng, Ding, Gao,
  Ge, et~al.]{bai2025qwen3}
Shuai Bai, Yuxuan Cai, Ruizhe Chen, Keqin Chen, Xionghui Chen, Zesen Cheng,
  Lianghao Deng, Wei Ding, Chang Gao, Chunjiang Ge, et~al.
\newblock Qwen3-vl technical report.
\newblock \emph{arXiv preprint arXiv:2511.21631}, 2025.

\bibitem[Bergmann et~al.(2019)Bergmann, Meinhardt, and Leal-Taixe]{Tracktor}
Philipp Bergmann, Tim Meinhardt, and Laura Leal-Taixe.
\newblock Tracking without bells and whistles.
\newblock In \emph{Proceedings of the IEEE/CVF international conference on
  computer vision}, pages 941--951, 2019.

\bibitem[Cao et~al.(2023)Cao, Pang, Weng, Khirodkar, and
  Kitani]{cao2023observation}
Jinkun Cao, Jiangmiao Pang, Xinshuo Weng, Rawal Khirodkar, and Kris Kitani.
\newblock Observation-centric sort: Rethinking sort for robust multi-object
  tracking.
\newblock In \emph{Proceedings of the IEEE/CVF conference on computer vision
  and pattern recognition}, pages 9686--9696, 2023.

\bibitem[Chen et~al.(2025)Chen, Yu, and Tao]{chen2025cross}
Sijia Chen, En Yu, and Wenbing Tao.
\newblock Cross-view referring multi-object tracking.
\newblock In \emph{Proceedings of the AAAI Conference on Artificial
  Intelligence}, pages 2204--2211, 2025.

\bibitem[Du et~al.(2024)Du, Lei, Zhao, and Su]{du2024ikun}
Yunhao Du, Cheng Lei, Zhicheng Zhao, and Fei Su.
\newblock ikun: Speak to trackers without retraining.
\newblock In \emph{Proceedings of the IEEE/CVF Conference on Computer Vision
  and Pattern Recognition}, pages 19135--19144, 2024.

\bibitem[Dunn(1974)]{dunn1974}
J.~C. Dunn.
\newblock Well-separated clusters and optimal fuzzy partitions.
\newblock \emph{Journal of Cybernetics}, 4\penalty0 (1):\penalty0 95--104,
  1974.

\bibitem[Feng et~al.(2021)Feng, Ablavsky, and Sclaroff]{Feng2021CityFlowNL}
Qi Feng, Vitaly Ablavsky, and Stan Sclaroff.
\newblock Cityflow-nl: Tracking and retrieval of vehicles at city scale by
  natural language descriptions.
\newblock \emph{arXiv preprint arXiv:2101.04741}, 2021.

\bibitem[He et~al.(2024)He, Jian, Lu, and Wang]{he2024visual}
Wenyan He, Yajun Jian, Yang Lu, and Hanzi Wang.
\newblock Visual-linguistic representation learning with deep cross-modality
  fusion for referring multi-object tracking.
\newblock In \emph{ICASSP 2024-2024 IEEE International Conference on Acoustics,
  Speech and Signal Processing (ICASSP)}, pages 6310--6314. IEEE, 2024.

\bibitem[Herzog et~al.(2024)Herzog, Gilg, Wolters, Teepe, and
  Rigoll]{herzog2024spatial}
Fabian Herzog, Johannes Gilg, Philipp Wolters, Torben Teepe, and Gerhard
  Rigoll.
\newblock Spatial-temporal multi-cuts for online multiple-camera vehicle
  tracking.
\newblock \emph{arXiv preprint arXiv:2410.02638}, 2024.

\bibitem[Hong et~al.(2023)Hong, Zheng, Chen, Cheng, Wang, Zhang, Wang, Yau,
  Lin, Zhou, et~al.]{metagpt}
Sirui Hong, Xiawu Zheng, Jonathan Chen, Yuheng Cheng, Jinlin Wang, Ceyao Zhang,
  Zili Wang, Steven Ka~Shing Yau, Zijuan Lin, Liyang Zhou, et~al.
\newblock Metagpt: Meta programming for multi-agent collaborative framework.
\newblock \emph{arXiv preprint arXiv:2308.00352}, 2023.

\bibitem[Huang et~al.(2025)Huang, Yao, Su, and Xu]{huang2025online}
Feijiang Huang, Jialong Yao, Chengyue Su, and Sheng Xu.
\newblock An online multi-camera multi-vehicle tracking using lightweight
  yolo11 and improved association strategies.
\newblock \emph{Journal of King Saud University Computer and Information
  Sciences}, 37\penalty0 (7):\penalty0 180, 2025.

\bibitem[Jocher and Qiu(2026)]{yolo26_ultralytics}
Glenn Jocher and Jing Qiu.
\newblock Ultralytics yolo26, 2026.

\bibitem[Li et~al.(2024)Li, Naseer, Cao, Zhu, Sun, Zhang, and
  Khan]{li2024multi}
Yuhao Li, Muzammal Naseer, Jiale Cao, Yu Zhu, Jinqiu Sun, Yanning Zhang, and
  Fahad~Shahbaz Khan.
\newblock Multi-granularity language-guided multi-object tracking.
\newblock \emph{arXiv preprint arXiv:2406.04844}, 2024.

\bibitem[Lin et~al.(2024)Lin, Chen, Peng, He, Li, Stiefelhagen, and
  Yang]{lin2024echotrack}
Jiacheng Lin, Jiajun Chen, Kunyu Peng, Xuan He, Zhiyong Li, Rainer
  Stiefelhagen, and Kailun Yang.
\newblock Echotrack: Auditory referring multi-object tracking for autonomous
  driving.
\newblock \emph{arXiv preprint arXiv:2402.18302}, 2024.

\bibitem[Luiten et~al.(2021)Luiten, Osep, Dendorfer, Torr, Geiger,
  Leal-Taix{\'e}, and Leibe]{luiten2021hota}
Jonathon Luiten, Aljosa Osep, Patrick Dendorfer, Philip Torr, Andreas Geiger,
  Laura Leal-Taix{\'e}, and Bastian Leibe.
\newblock Hota: A higher order metric for evaluating multi-object tracking.
\newblock \emph{International journal of computer vision}, 129:\penalty0
  548--578, 2021.

\bibitem[Ma et~al.(2024)Ma, Yang, Cui, Zhao, Su, Liu, and Wang]{ma2024mls}
Zeliang Ma, Song Yang, Zhe Cui, Zhicheng Zhao, Fei Su, Delong Liu, and Jingyu
  Wang.
\newblock Mls-track: Multilevel semantic interaction in rmot.
\newblock \emph{arXiv preprint arXiv:2404.12031}, 2024.

\bibitem[Nguyen et~al.(2024)Nguyen, Quach, Kitani, and Luu]{nguyen2024type}
Pha Nguyen, Kha~Gia Quach, Kris Kitani, and Khoa Luu.
\newblock Type-to-track: Retrieve any object via prompt-based tracking.
\newblock \emph{Advances in Neural Information Processing Systems}, 36, 2024.

\bibitem[Ren et~al.(2024)Ren, Jiang, Liu, Zeng, Liu, Gao, Huang, Ma, Jiang,
  Chen, et~al.]{ren2024grounding}
Tianhe Ren, Qing Jiang, Shilong Liu, Zhaoyang Zeng, Wenlong Liu, Han Gao,
  Hongjie Huang, Zhengyu Ma, Xiaoke Jiang, Yihao Chen, et~al.
\newblock Grounding dino 1.5: Advance the" edge" of open-set object detection.
\newblock \emph{arXiv preprint arXiv:2405.10300}, 2024.

\bibitem[Ristani et~al.(2016)Ristani, Solera, Zou, Cucchiara, and
  Tomasi]{ristani2016performance}
Ergys Ristani, Francesco Solera, Roger Zou, Rita Cucchiara, and Carlo Tomasi.
\newblock Performance measures and a data set for multi-target, multi-camera
  tracking.
\newblock In \emph{European conference on computer vision}, pages 17--35.
  Springer, 2016.

\bibitem[Shim et~al.(2023)Shim, Ko, Hwang, Jang, and Kim]{shim2023fast}
Kyujin Shim, Kangwook Ko, Jubi Hwang, Hyunsung Jang, and Changick Kim.
\newblock Fast online multi-target multi-camera tracking for vehicles.
\newblock \emph{Applied Intelligence}, 53:\penalty0 28994--29004, 2023.

\bibitem[Sz{\H{u}}cs et~al.(2023)Sz{\H{u}}cs, Borsodi, and
  Papp]{szucs2023hierarchical}
G{\'a}bor Sz{\H{u}}cs, Roland Borsodi, and D{\'a}niel Papp.
\newblock Multi-camera trajectory matching based on hierarchical clustering and
  constraints.
\newblock \emph{Multimedia Tools and Applications}, 2023.

\bibitem[Tang et~al.(2019)Tang, Naphade, Liu, Yang, Birchfield, Wang, Kumar,
  Anastasiu, and Hwang]{tang2019cityflow}
Zheng Tang, Milind Naphade, Ming-Yu Liu, Xiaodong Yang, Stan Birchfield, Shuo
  Wang, Ratnesh Kumar, David Anastasiu, and Jenq-Neng Hwang.
\newblock Cityflow: A city-scale benchmark for multi-target multi-camera
  vehicle tracking and re-identification.
\newblock In \emph{Proceedings of the IEEE/CVF Conference on Computer Vision
  and Pattern Recognition}, pages 8797--8806, 2019.

\bibitem[Tesfaye et~al.(2017)Tesfaye, Zemene, Prati, Pelillo, and
  Shah]{tesfaye2017}
Yonatan~Tariku Tesfaye, Eyasu Zemene, Andrea Prati, Marcello Pelillo, and
  Mubarak Shah.
\newblock Multi-target tracking in multiple non-overlapping cameras using
  constrained dominant sets, 2017.

\bibitem[Tran et~al.(2024)Tran, Le~Dinh, Nguyen, Phan, Nguyen, Luu, Adjeroh,
  Doretto, and Le]{tran2024z}
Kim Tran, Anh~Duy Le~Dinh, Tien-Phat Nguyen, Thinh Phan, Pha Nguyen, Khoa Luu,
  Donald Adjeroh, Gianfranco Doretto, and Ngan Le.
\newblock Z-gmot: Zero-shot generic multiple object tracking.
\newblock In \emph{Findings of the Association for Computational Linguistics:
  NAACL 2024}, pages 3468--3479, 2024.

\bibitem[Ullah and Alaya~Cheikh(2018)]{ullah2018directed}
Mohib Ullah and Faouzi Alaya~Cheikh.
\newblock A directed sparse graphical model for multi-target tracking.
\newblock In \emph{Proceedings of the IEEE Conference on Computer Vision and
  Pattern Recognition Workshops}, pages 1816--1823, 2018.

\bibitem[Wu et~al.(2023)Wu, Han, Wang, Dong, Zhang, and Shen]{wu2023referring}
Dongming Wu, Wencheng Han, Tiancai Wang, Xingping Dong, Xiangyu Zhang, and
  Jianbing Shen.
\newblock Referring multi-object tracking.
\newblock In \emph{Proceedings of the IEEE/CVF conference on computer vision
  and pattern recognition}, pages 14633--14642, 2023.

\bibitem[Xu et~al.(2022)Xu, Ban, Delorme, Gan, Rus, and
  Alameda-Pineda]{TransCenter}
Yihong Xu, Yutong Ban, Guillaume Delorme, Chuang Gan, Daniela Rus, and Xavier
  Alameda-Pineda.
\newblock Transcenter: Transformers with dense representations for
  multiple-object tracking, 2022.

\bibitem[Yao et~al.(2022)Yao, Duan, Xie, Chen, Wu, Xu, and Gao]{yao2022city}
Hui Yao, Zhizhao Duan, Zhen Xie, Jingbo Chen, Xi Wu, Duo Xu, and Yutao Gao.
\newblock City-scale multi-camera vehicle tracking based on
  space-time-appearance features.
\newblock In \emph{Proceedings of the IEEE/CVF conference on computer vision
  and pattern recognition}, pages 3310--3318, 2022.

\bibitem[Zhang et~al.(2024)Zhang, Wu, Han, and Dong]{zhang2024bootstrapping}
Yani Zhang, Dongming Wu, Wencheng Han, and Xingping Dong.
\newblock Bootstrapping referring multi-object tracking.
\newblock \emph{arXiv preprint arXiv:2406.05039}, 2024.

\bibitem[Zhao et~al.(2024)Zhao, Liu, He, Zhang, and Lee]{zhao2024real}
Tiancheng Zhao, Peng Liu, Xuan He, Lu Zhang, and Kyusong Lee.
\newblock Real-time transformer-based open-vocabulary detection with efficient
  fusion head.
\newblock \emph{arXiv preprint arXiv:2403.06892}, 2024.

\bibitem[Zhou et~al.(2021)Zhou, Yang, Qiao, and Xiang]{zhou2021mixstyle}
Kaiyang Zhou, Yongxin Yang, Yu Qiao, and Tao Xiang.
\newblock Domain generalization with mixstyle.
\newblock In \emph{International Conference on Learning Representations}, 2021.

\bibitem[Zhou et~al.(2020)Zhou, Koltun, and Kr{\"a}henb{\"u}hl]{CenterTrack}
Xingyi Zhou, Vladlen Koltun, and Philipp Kr{\"a}henb{\"u}hl.
\newblock Tracking objects as points.
\newblock In \emph{European conference on computer vision}, pages 474--490.
  Springer, 2020.

\end{thebibliography}
}


\end{document}


\title{Refer-CityFlow and M3Track: A Benchmark Dataset and Reference Framework for Language-Driven Multi-Target Multi-Camera Tracking\\
       \vspace{4pt}\large\textnormal{Supplementary Material}}
\author{
Phuong T. Nguyen$^{1}$\thanks{Equal contribution.}\quad
Hao Vo$^{2}$\footnotemark[1]\quad
Tam Bang$^{1}$\footnotemark[1]\quad
Hoang K. Tran$^{1}$\footnotemark[1]\\
Hoang-Minh Dau$^{3}$\quad
Mahshid Malazizi$^{1}$\quad
Tuan-Anh Hoang$^{3}$\quad
Mina Sartipi$^{1}$\\
Thanh-Nam Doan$^{4}$\quad
Ngan Le$^{2}$\quad
Hoang H. Nguyen$^{1}$\\[4pt]
$^{1}$University of Tennessee at Chattanooga \quad
$^{2}$University of Arkansas\\
$^{3}$RMIT International University Vietnam \quad
$^{4}$Independent Researcher\\
{\tt\small qlg645@mocs.utc.edu, huuhoang-nguyen@utc.edu}
}
\maketitle

\appendix
%
%

\noindent This supplement follows the order of the main paper: dataset details
(Sec.~\ref{sec:appendix_schema}--Sec.~\ref{sec:appendix_human_examples}),
evaluation splits and protocol
(Sec.~\ref{sec:appendix_scale}--Sec.~\ref{sec:appendix_protocol}), additional
experimental results
(Sec.~\ref{sec:appendix_results}--Sec.~\ref{sec:appendix_cityflow}),
reproducibility (Sec.~\ref{sec:appendix_geo}--Sec.~\ref{sec:appendix_prompts})
and a qualitative walkthrough (Sec.~\ref{sec:appendix_qualitative}), followed by
the dataset documentation. Summary statistics and the dataset-scale comparison
appear in Tab.~1 and Fig.~2 of the main paper and are not repeated here.

\section{Query Schema and Attribute Vocabulary}
\label{sec:appendix_schema}
Each query is represented by constraints over a fixed set of vehicle attributes
taken from the ground-truth database: \emph{color} (white, black, red, blue, gray/grey, silver,
brown, maroon, purple, green, tan, gold, yellow), \emph{vehicle type} (sedan,
SUV, truck, pickup, van, wagon, hatchback, coupe, jeep, minivan; this list seeds
the Requirement Analyzer (Sec.~\ref{sec:appendix_prompts}), whose schema stores
vehicle type as free text, so types outside the seed list, e.g., buses referenced in
real-user queries, remain representable), \emph{camera ID}, \emph{direction}
(e.g., east, west, north, south; annotation uses world-frame cardinal
directions, while the parsing schema additionally accepts the screen-frame terms
left/right/top/bottom as they occur in user phrasing), \emph{zone} (the lettered
entry/exit/transition regions keyed to each camera), \emph{action} (e.g., go
straight, turn, stop), \emph{lane}, \emph{relation} (interactions with nearby
objects), and \emph{time range}. A query is represented as a set of constraints
over these fields together with an optional list of reappearance cameras; the six
query formats (QF-1--QF-6) in the main paper correspond to which subsets of these
fields are constrained.

\paragraph{Primary and auxiliary cameras.}
Every AI-generated query names one \emph{primary camera}, in which the target
must appear within the stated time window, together with one or more
\emph{auxiliary} cameras in which it must reappear. A real-user query names a
primary camera only when the participant supplies one; where the resolved form
leaves it open, the constraint is simply absent. The distinction matters for the protocol: the time
window is applied strictly to the primary camera and relaxed on auxiliary
cameras (Sec.~5.1 of the main paper), and the reference associator uses the
primary camera as the trajectory anchor (Sec.~4.4).

\subsection*{Spatial zoning}
\begin{figure}[!tb]
\centering
\includegraphics[width=\linewidth]{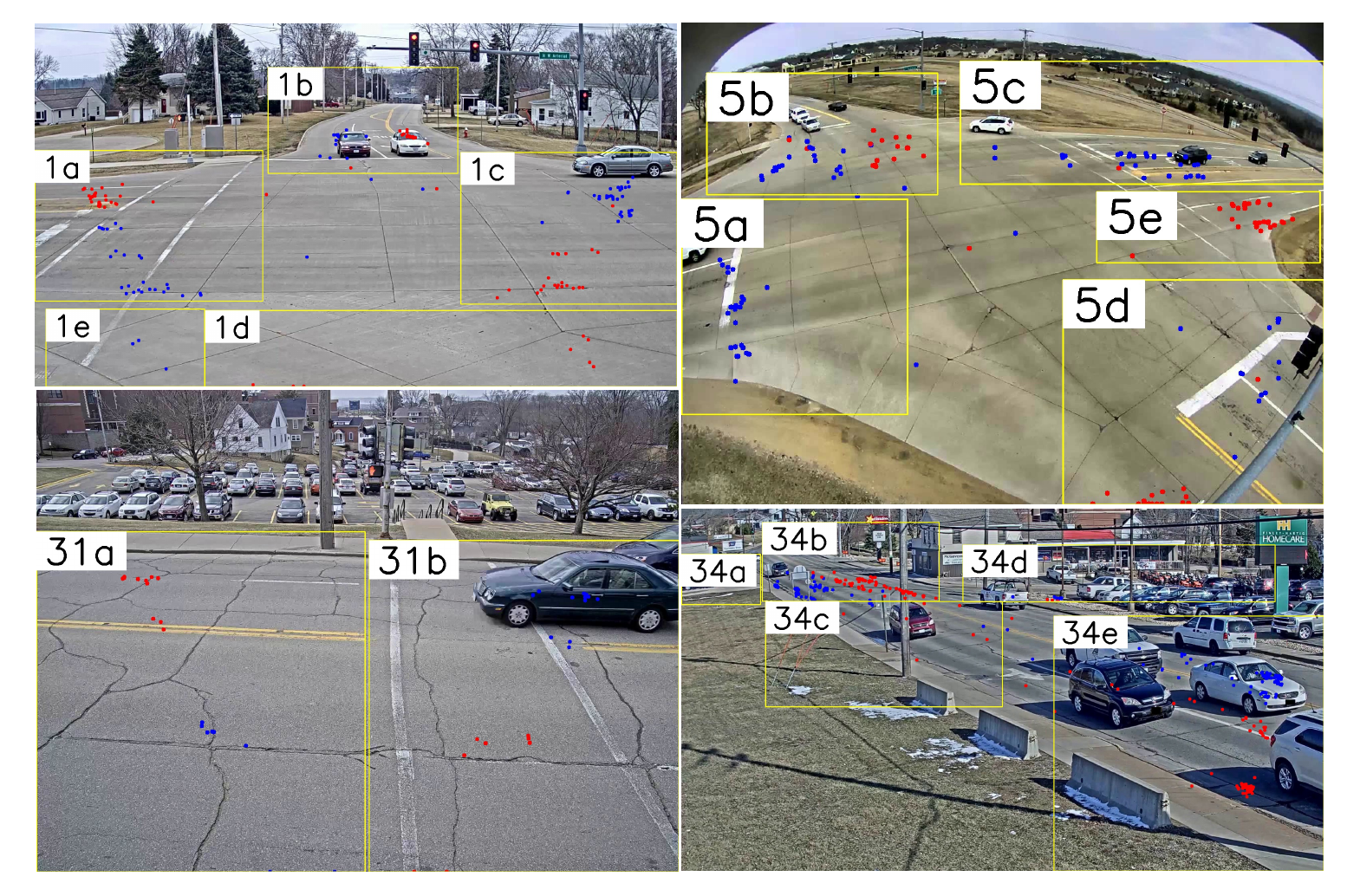}
\caption{\textbf{Per-camera spatial zoning.} Each camera view is annotated with
lettered zones keyed to the camera ID (entry, exit, and transition regions),
enabling structured spatial reasoning within a view and queries that name the exit zone of one camera together with the entry zone of an
adjacent camera.}
\label{fig:zoning}
\end{figure}

A key feature of \ourdata is its structured spatial annotation. Within each camera view we define \emph{zones}: entry, exit, and transition regions, which segment the road surface into semantically meaningful areas
(Fig.~\ref{fig:zoning}). Zones are labeled per camera with identifiers of the
form ``1a, 1b, 1c'' keyed to the camera ID, so that a query such as ``a vehicle
leaving through the exit of camera~1 and entering camera~2'' can be grounded and
verified against the ground truth. The zone annotations provide two capabilities that bounding-box annotations
alone do not: structured spatial reasoning within a view, and
cross-camera hand-off, where the exit zone of one camera and the entry zone of
an adjacent camera can be named in the same query. Combined with lane labels, zones let the
benchmark pose queries about direction of travel, turning behavior, and
cross-camera transitions that require reasoning over the road topology. The zone labels used by the queries are the ones
carried by the annotations; Fig.~\ref{fig:zoning} shows representative views.

\section{AI-Generated Query Examples}
\label{sec:appendix_ai_examples}
Every query is generated from the ground-truth database and validated by
executing the corresponding database query and recording the matching vehicle
and detection counts (Sec.~\ref{sec:appendix_genprompts}).
Tab.~\ref{tab:query_example_full} lists ten examples per format.

\begin{table*}[htb]
\footnotesize
\caption{\textbf{Examples of AI-generated queries}, ten per format, reproduced
\emph{verbatim} from the released query files. QF: query format; Scenario: the
CityFlow scenario the query targets. The examples retain the template-driven
grammar of the generator (e.g., singular nouns after ``all'', ``that moving''),
which we keep unedited so that they match the released data exactly.}
\resizebox{2\columnwidth}{!}{
\begin{tabular}{|l|c|l|}
\hline
QF & Scenario   & Queries                                                                                                                                                        \\ \hline
  1 & 4 & Track all white truck that appear in camera 16, and also appear at camera 17 \\ \hline
  1 & 4 & Track all black minivan that appear in camera 32 from 25.1 second, and also appear at camera 31 \\
  \hline
  1 & 4 & Track all white truck that appear in camera 37, and also appear at camera 36 \\ \hline
  1 & 2 & Track all white suv that appear in camera 8, and also appear at camera 9 \\ \hline
  1 & 5 & Track all black suv that appear in camera 33 from 312.1 second, and also appear at camera 17 \\ \hline
  1 & 1 & Track all blue truck that appear in camera 1 from 193.5 second to 195.3 second, and also appear at
  camera 2 \\ \hline
  1 & 2 & Track all white truck that appear in camera 7 from 124.1 second, and also appear at camera 6 \\ \hline
  1 & 1 & Track all red sedan that appear in camera 5 from 207.3 second, and also appear at camera 3 \\ \hline
  1 & 5 & Track all maroon sedan that appear in camera 33 from 234.1 second, and also appear at camera 35 \\
  \hline
  1 & 5 & Track all white truck that appear in camera 29, and also appear at camera 28,27 \\ \hline

  2 & 1 & Track all brown truck that moving from east to north in camera 4 from 33.8 second, and also appear at
  camera 1 \\ \hline
  2 & 5 & Track all gray suv that moving from southwest to east in camera 23 from 24.2 second, and also appear at
  camera 24 \\ \hline
  2 & 5 & Track all black minivan that moving from northeast to southeast in camera 28 from 253.4 second to 273.6
  second, and also appear at camera 34,29 \\ \hline
  2 & 5 & Track all black suv that moving from northeast to southeast in camera 35, and also appear at camera 36
  \\ \hline
  2 & 5 & Track all black suv that moving from south to northeast in camera 29 from 194.0 second to 226.2 second,
  and also appear at camera 35,34 \\ \hline
  2 & 3 & Track all black suv that moving from east to west in camera 13 from 212.9 second to 221.4 second, and
  also appear at camera 10 \\ \hline
  2 & 1 & Track all black truck that moving from east to north in camera 2, and also appear at camera 1,3 \\
  \hline
  2 & 1 & Track all white truck that moving from middle to north in camera 2 from 0.1 second, and also appear at
  camera 5 \\ \hline
  2 & 1 & Track all white van that moving from middle to west in camera 1 from 51.8 second to 61.4 second, and
  also appear at camera 5 \\ \hline
  2 & 5 & Track all silver sedan that moving from south to east in camera 29 from 71.0 second, and also appear at
  camera 33,34 \\ \hline

  3 & 5 & Track all the black suv that moving from zone 29b to zone 29c in camera 29 from 300.6 second to 301.3
  second, and also appear at camera 18,22,34 \\ \hline
  3 & 5 & Track all the maroon suv that moving from zone 25b to zone 25a in camera 25 from 221.4 second to 225.7
  second, and also appear at camera 20,22 \\ \hline
  3 & 5 & Track all the green truck that moving from zone 33c to zone 33a in camera 33 from 108.5 second to 116.5
  second, and also appear at camera 19,20,23 \\ \hline
  3 & 5 & Track all the black sedan that moving from zone 22d to zone 22b in camera 22 from 15.8 second to 25.3
  second, and also appear at camera 16,17,18 \\ \hline
  3 & 1 & Track all the blue suv that moving from zone 1c to zone 1a in camera 1 from 40.2 second to 41.2 second,
  and also appear at camera 2,3,4 \\ \hline
  3 & 5 & Track all the silver sedan that moving from zone 33c to zone 33a in camera 33 from 123.9 second to 133.3
   second, and also appear at camera 34,35 \\ \hline
  3 & 5 & Track all the red suv that moving from zone 29a to zone 29c in camera 29 from 214.5 second to 250.1
  second, and also appear at camera 36 \\ \hline
  3 & 5 & Track all the white suv that moving from zone 22b to zone 22d in camera 22 from 143.8 second to 154.4
  second, and also appear at camera 19,20,25 \\ \hline
  3 & 4 & Track all the white truck that moving from zone 17c to zone 17b in camera 17 from 13.2 second to 15.9
  second, and also appear at camera 16 \\ \hline
  3 & 5 & Track all the maroon suv that moving from zone 25b to zone 25a in camera 25 from 221.4 second to 225.7
  second, and also appear at camera 17 \\ \hline

  4 & 1 & Track all white suv that goes straight in camera 1 from 92.0 second to 95.8 second, and also appear at
  camera 4,2 \\ \hline
  4 & 2 & Track all black suv that goes straight in camera 6 from 17.6 second to 20.0 second, and also appear at
  camera 8,9 \\ \hline
  4 & 5 & Track all black wagon that goes straight in camera 36 from 308.8 second to 339.5 second, and also appear
   at camera 35,33 \\ \hline
  4 & 1 & Track all blue truck that goes straight in camera 4 from 0.1 second to 4.7 second, and also appear at
  camera 1,2 \\ \hline
  4 & 2 & Track all white sedan that turns left in camera 8, and also appear at camera 7 \\ \hline
  4 & 2 & Track all gray sedan that goes straight in camera 8 from 110.1 second to 111.2 second, and also appear
  at camera 9 \\ \hline
  4 & 2 & Track all black truck that goes straight in camera 6 from 19.4 second, and also appear at camera 9,7 \\
  \hline
  4 & 2 & Track all black sedan that goes straight in camera 9 from 105.3 second to 109.0 second, and also appear
  at camera 6 \\ \hline
  4 & 1 & Track all white minivan that turns right in camera 3 from 56.8 second, and also appear at camera 5 \\
  \hline
  4 & 2 & Track all white truck that goes straight in camera 6 from 105.7 second to 108.5 second, and also appear
  at camera 8 \\ \hline

  5 & 2 & Track all black suv that are behind a black vehicle in camera 6 from 31.5 second to 31.9 second, and
  also appear at camera 9 \\ \hline
  5 & 1 & Track all blue suv that are alongside a black vehicle in camera 4, and also appear at camera 1,3,5 \\
  \hline
  5 & 5 & Track all blue suv that are alongside a red vehicle in camera 18 from 375.6 second to 376.8 second, and
  also appear at camera 24,23 \\ \hline
  5 & 5 & Track all silver truck that are behind a white vehicle in camera 36 from 13.1 second, and also appear at
   camera 35 \\ \hline
  5 & 4 & Track all white truck that are in front of a gray vehicle in camera 40 from 11.5 second to 12.8 second,
  and also appear at camera 39 \\ \hline
  5 & 2 & Track all white sedan that are in front of a black vehicle in camera 9, and also appear at camera 7,8 \\
   \hline
  5 & 2 & Track all gray minivan that are behind a white vehicle in camera 9 from 184.2 second, and also appear at
   camera 8,7 \\ \hline
  5 & 2 & Track all black sedan that are behind a black vehicle in camera 9 from 8.2 second, and also appear at
  camera 6 \\ \hline
  5 & 2 & Track all black sedan that are alongside a gray vehicle in camera 6 from 8.7 second, and also appear at
  camera 7 \\ \hline
  5 & 3 & Track all white sedan that are alongside a white vehicle in camera 10 from 118.1 second, and also
  appear at camera 12,11 \\ \hline

  6 & 5 & Track all the silver sedan that moving in lane number 4 in camera 22 from 35.7 second to 42.5 second,
  and also appear at camera 19,27,28 \\ \hline
  6 & 5 & Track all the gray truck that moving in lane number 1 in camera 17 from 247.2 second to 251.5 second,
  and also appear at camera 16,19,20 \\ \hline
  6 & 5 & Track all the gray suv that moving in lane number 3 in camera 26 from 57.3 second to 59.3 second, and
  also appear at camera 18,29,35 \\ \hline
  6 & 2 & Track all the black truck that moving in lane number 2 in camera 6 from 19.4 second to 22.7 second, and
  also appear at camera 7 \\ \hline
  6 & 4 & Track all the white truck that moving in lane number 0 in camera 35 from 10.8 second to 14.8 second, and
   also appear at camera 23,26,30 \\ \hline
  6 & 5 & Track all the silver sedan that moving in lane number 13 in camera 33 from 170.9 second to 172.4 second,
   and also appear at camera 22,25,26 \\ \hline
  6 & 5 & Track all the white suv that moving in lane number 12 in camera 33 from 264.8 second to 265.4 second,
  and also appear at camera 16,25,28 \\ \hline
  6 & 1 & Track all the white suv that moving in lane number 3 in camera 3 from 19.2 second to 87.7 second, and
  also appear at camera 5 \\ \hline
  6 & 5 & Track all the blue suv that moving in lane number 0 in camera 26 from 215.3 second to 217.0 second, and
  also appear at camera 16,17,24 \\ \hline
  6 & 5 & Track all the gray suv that moving in lane number 3 in camera 25 from 274.7 second to 278.6 second, and
  also appear at camera 21,26,27 \\ \hline
\end{tabular}}

\label{tab:query_example_full}
\end{table*}

Every query in this set specifies both a vehicle color and type, along with at
least one reappearance camera, so answering it requires both attribute grounding and
cross-camera association.

\section{Query Generation Prompts}
\label{sec:appendix_genprompts}
\paragraph{Generation pipeline.}
The 6{,}000 AI-generated queries are produced in four steps. (i)~For each
scenario, all vehicles are retrieved from the ground-truth database. (ii)~For
each vehicle, the attributes required for the selected query format (color, vehicle
type, camera, direction, zone, action, relation, lane, and time range) are
extracted, together with the other cameras in which that vehicle appears.
(iii)~The attributes are passed to the language model, which produces three temporal variants per vehicle (no time filter, start-time filter, and time-range filter), each requiring the vehicle to reappear in one or more
further cameras drawn at random from the same scenario. (iv)~Every generated query is
translated into an SQL query and executed against the database; the returned
vehicle and detection counts are recorded, and a query is kept only if it
returns at least one matching trajectory, so every released query has a
non-empty target set.

Fig.~\ref{fig:query_generation} summarizes the flow.

\begin{figure}[!tb]
  \centering
  \begin{tikzpicture}[node distance=0.5cm, scale=0.9, transform shape]
      \node (start)   [startstop] {Start};
      \node (db)      [process,  below=of start]  {Connect to ground-truth DB};
      \node (loop1)   [process,  below=of db]     {For each scenario $s$};
      \node (getveh)  [process,  below=of loop1]  {Retrieve and shuffle vehicles};
      \node (loop2)   [process,  below=of getveh] {For each vehicle $v$};
      \node (dec1)    [decision, below=of loop2]  {Valid segments \&\\other cameras?};
      \node (combo)   [process,  below=of dec1]   {Build attribute combinations\\(3 time variations)};
      \node (openai)  [io,       below=of combo]  {LLM $\rightarrow$ NL query};
      \node (exec)    [process,  below=of openai] {Build \& execute SQL\\on ground-truth DB};
      \node (out)     [io,       below=of exec]   {Output record};
      \node (dec3)    [decision, below=of out]    {Any matching\\trajectory?};
      \node (dec4)    [decision, below=of dec3]   {More scenarios?};
      \node (stop)    [startstop, below=of dec4]  {End};

      \draw[arrow] (start)  -- (db);
      \draw[arrow] (db)     -- (loop1);
      \draw[arrow] (loop1)  -- (getveh);
      \draw[arrow] (getveh) -- (loop2);
      \draw[arrow] (loop2)  -- (dec1);
      \draw[arrow] (dec1)   -- node[right, font=\scriptsize] {Yes} (combo);
      \draw[arrow] (dec1.west) -- ++(-1.0,0) node[midway, above, font=\scriptsize] {No} |- (loop2.west);
      \draw[arrow] (combo)  -- (openai);
      \draw[arrow] (openai) -- (exec);
      \draw[arrow] (exec)   -- (out);
      \draw[arrow] (out)    -- (dec3);
      \draw[arrow] (dec3)   -- node[right, font=\scriptsize] {Yes} (dec4);
      \draw[arrow] (dec3.east) -- ++(1.0,0) node[midway, above, font=\scriptsize] {No} |- (loop2.east);
      \draw[arrow] (dec4)   -- node[right, font=\scriptsize] {No} (stop);
      \draw[arrow] (dec4.east) -- ++(1.6,0) node[midway, above, font=\scriptsize] {Yes} |- (loop1.east);
  \end{tikzpicture}
  \caption{\textbf{Query generation pipeline.} Vehicles are drawn from the
  ground-truth database, their attributes are combined into three temporal
  variants, an LLM renders each combination as a natural-language query, and the
  query is executed against the database; queries that return no matching trajectory are discarded and the vehicle is
  resampled.}
  \label{fig:query_generation}
\end{figure}

\paragraph{Query templates.}
All six formats share one sentence template ( \textit{Track all the
$\langle$color$\rangle$ $\langle$vehicle\_type$\rangle$ that
$\langle$format-specific constraint$\rangle$ in camera
$\langle$cameraID$\rangle$ $\langle$time$\rangle$, and also appear at camera
$\langle$cameraIDs$\rangle$}), and differ only in the constraint that defines
the format. The slots are filled from the annotation of the selected vehicle, so their
values are taken directly from the recorded annotations:
$\langle$color$\rangle$ and $\langle$vehicle\_type$\rangle$ come from the
appearance attributes, $\langle$cameraID$\rangle$ is a camera in which the
vehicle is observed (the primary camera), $\langle$time$\rangle$ is its observed
segment in that camera, and $\langle$cameraIDs$\rangle$ are one to three further
cameras in which the same global identity also appears. The format-specific
constraint is:
\begin{itemize}
\item \textbf{QF-1:} \emph{appear}, no additional slot.
\item \textbf{QF-2:} \emph{moving from $\langle$direction1$\rangle$ to
$\langle$direction2$\rangle$}, the annotated entry and exit directions, given as
world-frame cardinal terms (north, east, southwest, \ldots) or \emph{middle} for
a central approach.
\item \textbf{QF-3:} \emph{moving from zone $\langle$zoneID1$\rangle$ to zone
$\langle$zoneID2$\rangle$}, the annotated entry and exit zones of the primary
camera.
\item \textbf{QF-4:} \emph{$\langle$action$\rangle$}, the annotated action:
goes straight, turns left, turns right, or stops.
\item \textbf{QF-5:} \emph{$\langle$relation$\rangle$ a
$\langle$color$\rangle$ vehicle}, with the relation drawn from \{are behind, are
in front of, are alongside\} and the neighboring object described by its color.
\item \textbf{QF-6:} \emph{moving in lane number $\langle$laneID$\rangle$}, the
annotated numeric lane index of the primary camera.
\end{itemize}
The generation prompts will be released verbatim with the code.

\section{Real-User Query Refinement Examples}
\label{sec:appendix_human_examples}
Real-user queries were collected through a lightweight web frontend built with
Gradio, in which participants type a request in free form and answer the
clarification questions the agent raises; Fig.~3 of the main paper gives the
resulting refinement workflow. Real-user queries are often underspecified, and
the system resolves them into an executable form through this interactive
clarification.
Tab.~\ref{tab:refinement_examples} lists representative before/after pairs.

\begin{table*}[t]
\caption{\textbf{Real-user queries before and after interactive refinement},
reproduced \emph{verbatim}. Camera IDs of the form C0xx are reproduced as typed
by the participant. Refinement normalizes queries to set-valued semantics, e.g.,
``the red sedan'' resolves to ``all red sedans'' matching the constraints.}
\label{tab:refinement_examples}
\centering
\footnotesize
\setlength{\tabcolsep}{4pt}
\renewcommand{\arraystretch}{0.95}
\begin{tabularx}{\textwidth}{
    >{\raggedright\arraybackslash}p{0.05\textwidth}
    >{\raggedright\arraybackslash}p{0.28\textwidth}
    >{\raggedright\arraybackslash}X
}
\toprule
No. & Before refinement (Initial Raw Query) & After refinement \\
\midrule

1 & Track the red sedan in C025 & Track all red sedans that appear in camera C025 from 0 to 7 seconds and also appear in camera C022 \\
2 & track white suv goes straight at intersection in c8 & Track all white SUVs that go straight in camera 8 \\
3 & Track the bus turns left at intersection in cam28 & Track all green buses that turn left in camera 28 \\
4 & I want to track the blue sedan goes straight in cam 10, 11, 14 & Track all blue sedans that go straight in camera 10 and also appear in cameras 11 and 14 \\
5 & show me all information about the fedex van & Track all vans that appear \\
6 & track all trucks go straight at the intersection in camera 29 & Track all white trucks that go straight in camera 29 \\
7 & Track container truck goes straight in camera 6 & Track all white trucks that go straight in camera 6 \\
8 & Track red car & Track all red sedans that appear in camera C004 from 9 to 33 seconds and also appear in cameras C001--C040 \\
9 & tracking the white van goes straight from cam 31 to 33 & Track all white vans that go straight in camera 31 and also appear in camera 33 \\
10 & track red sedan in cam 22 and 25 goes straight & Track all red sedans that go straight in camera 22 and also appear in camera 25 \\
11 & all trucks turn right in cam 2 and cam 5 & Track all trucks that turn right in camera 2 and also appear in camera 5 \\
12 & I want white van appear in cam 31 to cam 33 & Track all white vans that appear in camera 31 and also appear in cameras 32 and 33 \\
13 & show green truck presents at the intersection & Track all green vehicles moving in the intersection \\
14 & I need a maroon sedan in the right lane & Track all maroon sedans moving in the right lane in camera 2 \\
15 & track the red sedan goes straight first appears on cam 25 & Track all red sedans that go straight in camera 25 \\
16 & track green bus in cam 28, 33, 34 & Track all green buses that appear in camera 28 and also appear in cameras 33 and 34 \\
17 & I'm wanting a white minivan that goes straight under the traffic light & Track all white minivans that go straight \\
18 & White GMC SUV moving on the road & Track all white SUVs that appear in camera C021 from 0 to 31 seconds and also appear in camera C028 \\
19 & I'm tracking a silver truck that stops in the single right lane & Track all silver trucks moving in the single right lane \\
20 & track red sedan goes straight in camera 22 and 23 & Track all red sedans that go straight in camera 22 and also appear in camera 23 \\
21 & track all silver car stop at the red light & Track all silver vehicles that stop at the traffic light in camera 3 \\
22 & I need a black suvedan in the middle lane & Track all black SUVs moving in the middle lane \\
23 & track the black sedan turns left in camera 10 & Track all black sedans that turn left in camera 10 \\
24 & track all green trucks go straight at the intersection & Track all green trucks moving in the intersection \\
25 & blue truck goes straight at the intersection & Track all blue trucks that go straight \\
26 & I want a gray sedan that goes straight in the left lane & Track all gray sedans moving in the left lane \\
27 & Track white trucks go straight & Track all white trucks that go straight \\
28 & show all white cars stop at the red light & Track all white vehicles that stop at the traffic light \\
29 & I'm finding a black suv that turns right & Track all black SUVs that turn right \\
30 & I want a white container truck that goes straight in the right lane & Track all white trucks moving in the right lane \\

\bottomrule
\end{tabularx}
\end{table*}

\section{Evaluation Splits and Release Scope}
\label{sec:appendix_scale}

\paragraph{Splits and use.}
Queries are grouped by source (AI-generated vs.\ real-user) and, for the
AI-generated set, by query format. Grouping queries this way supports
capability-specific analysis across language grounding, cross-camera
association, and spatio-temporal reasoning rather than relying only on a single
aggregate score. No portion of the benchmark is designated for training: all reference
results in the main paper are zero-shot, and the full benchmark is intended as
an evaluation set. Methods that require supervision should train on external
data and report the split they used.

\paragraph{Release scope.}
The release covers the complete benchmark (all 40 cameras, all annotated trajectories, and both query sets); the evaluation protocol and reference implementation will follow in the project repository. Scenario~02 is used for the reference experiments
of the main paper because the reference cross-camera associator relies on
overlapping fields of view (Sec.~\ref{sec:appendix_geo}); it is a property of
that associator, not a restriction of the benchmark or of the release.
Camera configurations that include non-overlapping views require an associator
that does not rely exclusively on overlap.

\section{Evaluation Protocol Details}
\label{sec:appendix_protocol}

\paragraph{IDF1.}
IDF1~\cite{ristani2016performance} is the identity-based F1 score, the ratio of
correctly identified detections to the average number of ground-truth and
predicted detections:
\begin{equation}
\mathrm{IDF1}=\frac{2\,\mathrm{IDTP}}{2\,\mathrm{IDTP}+\mathrm{IDFP}+\mathrm{IDFN}},
\label{eq:idf1}
\end{equation}
where IDTP, IDFP and IDFN count correct, false-positive and false-negative
\emph{identity} assignments under a one-to-one matching between predicted and
ground-truth identities.

\paragraph{HOTA, DetA and AssA.}
HOTA~\cite{luiten2021hota} measures the agreement between predicted and
ground-truth trajectories while penalizing unmatched detections, giving equal
weight to detection and association quality. At a localization threshold
$\alpha$,
\begin{align}
\mathrm{DetA}_\alpha &= \frac{|\mathrm{TP}|}{|\mathrm{TP}|+|\mathrm{FN}|+|\mathrm{FP}|}, \\
\mathcal{A}(c) &= \frac{|\mathrm{TPA}(c)|}{|\mathrm{TPA}(c)|+|\mathrm{FNA}(c)|+|\mathrm{FPA}(c)|}, \\
\mathrm{AssA}_\alpha &= \frac{1}{|\mathrm{TP}|}\sum_{c\in\mathrm{TP}} \mathcal{A}(c), \\
\mathrm{HOTA}_\alpha &= \sqrt{\frac{\sum_{c\in\mathrm{TP}}\mathcal{A}(c)}{|\mathrm{TP}|+|\mathrm{FN}|+|\mathrm{FP}|}}
 = \sqrt{\mathrm{DetA}_\alpha\cdot\mathrm{AssA}_\alpha},
\end{align}
where $\mathcal{A}(c)$ is the association score of a matched detection $c$, and
TPA, FNA and FPA are its true-positive, false-negative and false-positive
associations. TP, FN and FP follow the definitions used for
MOTA~\cite{bernardin2008evaluating}. The reported HOTA integrates
$\mathrm{HOTA}_\alpha$ over the standard range of $\alpha$. As in standard
referring-tracking evaluation, a visible vehicle that is not referred to by the
query counts as a false positive if it is returned.

\paragraph{L2 distance (counting error).}
We use the L2 distance (Euclidean distance) to evaluate counting accuracy. Given
a query, the model or the human annotator predicts a number of matched target
objects $\hat{n}$, which is compared against the ground-truth count $n$. For a
single query,
\begin{equation}
d_{\mathrm{L2}}(n,\hat{n}) = \sqrt{(n-\hat{n})^2} = \left|n-\hat{n}\right|,
\end{equation}
and over a set of $Q$ queries we report the mean L2 distance as the overall
counting error,
\begin{equation}
\overline{d}_{\mathrm{L2}} = \frac{1}{Q}\sum_{i=1}^{Q}\left|n_i-\hat{n}_i\right|,
\end{equation}
where $n_i$ and $\hat{n}_i$ are the ground-truth and predicted counts for the
$i$-th query. A lower mean L2 distance indicates that the system more accurately
identifies the number of target objects satisfying the query conditions. Because the counts are scalars, this quantity coincides with the mean absolute
error, which the main paper reports as the mean L1 target-count error.

\paragraph{Scope of the identity matching.}
IDF1 depends on the scope over which the one-to-one matching is computed, and we
state it explicitly at each use. In Tabs.~5 and~6 of the main paper the
one-to-one matching is performed \emph{within each (query, camera) pair}, and
the resulting IDTP/IDFP/IDFN counts are pooled across all pairs before IDF1 is
computed; this follows the per-sequence procedure of TrackEval~\cite{trackeval}. Because the ground truth of a (query, camera)
pair contains only the trajectories that satisfy the query, a system that fails
to link identities across views cannot recover the correct target set in the
auxiliary cameras, and is penalized through the composition of the returned set
rather than through the identity labels themselves. Tab.~\ref{tab:cross_cam_idf1}
instead reports IDF1 with the matching performed over \emph{global identities
spanning all cameras}, which isolates the effect of cross-camera association;
the two scopes are not directly comparable.\\
Under the per-pair scope, a correctly returned vehicle that receives different
labels in different cameras is not penalized by IDF1; this kind of cross-camera
fragmentation is measured by the global-identity IDF1 and identity switches in
Tab.~\ref{tab:cross_cam_idf1}.

\paragraph{Baseline scoring.}
TransRMOT and TempRMOT emit per-camera output and no cross-camera identities.
Their numbers indicate single-camera referring performance; they are not a matched comparison with systems that return cross-camera trajectories.

\paragraph{Submission format.}
A system is evaluated from one plain-text file per (query, camera) pair, named
after the sequence \texttt{S\{ss\}-camera\_\{cam\}-query\_\{qid\}} with a
two-digit scenario index (e.g., \texttt{S02-camera\_6-query\_128}), in
MOTChallenge format: one line per detection,
\begin{center}
\texttt{frame, global\_id, bb\_left, bb\_top, bb\_width, bb\_height, 1, -1, -1, -1}
\end{center}
with \texttt{frame} the frame index within that camera,
\texttt{global\_id} the identity assigned by the system (consistent across
cameras for the same physical vehicle), and the bounding box in pixel
coordinates. Predictions are matched to ground truth by IoU, using the standard
HOTA threshold range and a fixed threshold of $0.5$ for IDF1. A (query, camera)
pair for which the system returns nothing is scored as an empty file rather than
skipped, so its ground-truth trajectories count as false negatives.

\section{Additional Results}
\label{sec:appendix_results}

\subsection{Cross-Camera Module Ablation}
\label{sec:appendix_mcm}
Tab.~\ref{tab:cross_cam_idf1} isolates the Multi-Camera Module (MCM) on
CityFlow~S02 under the simplified ``track all vehicles'' setting, i.e., without
language constraints, so that the comparison isolates the effect of the
association module. IDF1 here is computed over global identities spanning all cameras
(Sec.~\ref{sec:appendix_protocol}). The MCM raises this IDF1 by $+$7.3
(YOLOv5+SVM) and $+$10.9 (YOLO26-m+SVM) and approximately halves identity
switches, while IDF1 computed within each camera remains stable, consistent with
the improvement arising from cross-camera identity association so the gain comes from cross-camera association. Without the MCM each camera produces its own local track IDs, so a
vehicle crossing several views is counted as several separate identities; the
hierarchical clustering and association steps of the MCM link those tracklets
into one globally consistent trajectory.

\begin{table}[t]
\caption{\textbf{IDF1 over global identities, before and after the Multi-Camera
Module (MCM)}, CityFlow S02, ``track all vehicles'' setting. IDSW: identity
switches in global association.}
\centering
\resizebox{\columnwidth}{!}{
\begin{tabular}{l|cc|cc|cc}
\toprule
& \multicolumn{2}{c|}{\textbf{Before MCM}} & \multicolumn{2}{c|}{\textbf{After MCM}} & \multicolumn{2}{c}{\textbf{$\Delta$}} \\
\textbf{Detector} & IDF1 & IDSW & IDF1 & IDSW & IDF1 & IDSW \\
\midrule
YOLOv5+SVM    & 60.5 & 2136 & \textbf{67.8} & \textbf{1187} & $+$7.3  & $-$949 \\
YOLO26-m+SVM  & 55.7 & 1941 & \textbf{66.6} & \textbf{972}  & $+$10.9 & $-$969 \\
\bottomrule
\end{tabular}}
\label{tab:cross_cam_idf1}
\end{table}

\subsection{Human vs.\ Automatic Evaluation}
\label{sec:appendix_human}
To provide a small-scale human reference for the counting task, we conducted a
manual evaluation on the real-user query set, which includes the
kinds of underspecification common in user-authored requests. We recruited
three annotators and assigned each of them 10 queries from Scenario~02. The
annotators watched the multi-camera videos and counted, by hand, the target
vehicles matching each query description. For the automatic systems the target
count is obtained by matching predicted tracklets to ground-truth identities via
IoU overlap and aggregating true-positive identity pairs: a predicted track $t$
counts as a match if at least one of its boxes overlaps a ground-truth box above
an IoU threshold $\tau=0.5$,
\begin{equation}
\text{match}(t)=\mathbb{1}\!\left[\max_i \text{IoU}(\mathbf{b}_t^{i},\mathbf{b}_{gt}^{i})>\tau\right],
\end{equation}
and the number of matched tracks gives the predicted count $\hat{n}$. The L1
error is the absolute difference $|n-\hat{n}|$ between the ground-truth count
$n$ and the predicted (or human-reported) count, averaged over queries.

\paragraph{Human protocol.}
For each assigned query the annotator started a stopwatch, read the query to
identify the target attributes and constraints, then examined the videos of the
relevant cameras and searched manually for vehicles satisfying the conditions,
noting candidates and comparing them across views. After reviewing all relevant
videos the annotator finalized the target set, recorded the count, and stopped
the stopwatch. Completion times were summed across queries to obtain the
total human evaluation time reported in Tab.~\ref{tab:res_human_perf_query}.

\paragraph{Result.}
With the Grounding-DINO backbone, \ourmodel reaches an L1 error of 30.15 against
a human reference of 30.50, at 948\,s of total processing versus 5{,}268\,s for
the annotators. TransRMOT and TempRMOT run faster (627\,s and 724\,s) but with
higher L1 error (33.45 and 33.00).

\begin{table}[t]
\caption{\textbf{Human vs.\ automatic evaluation} on \ourdata.}
\setlength{\tabcolsep}{9pt}
\centering
\resizebox{0.99\columnwidth}{!}{
\begin{tabular}{l|c|cc}
\toprule
\multicolumn{1}{c|}{\textbf{Model}} &
\multicolumn{1}{c|}{\textbf{Detection Backbone}} &
\multicolumn{1}{c}{\textbf{L1$\downarrow$}} &
\multicolumn{1}{c}{\textbf{Total time (s)$\downarrow$}} \\ \toprule
TransRMOT \cite{wu2023referring} & Def.\ DETR & 33.45 & \textbf{627} \\
TempRMOT \cite{zhang2024bootstrapping} & Def.\ DETR & 33.00 & 724 \\
\textbf{\ourmodel (ours)} & Grounding-DINO & \textbf{30.15} & 948 \\ \midrule
Human & -- & 30.50 & 5,268 \\ \bottomrule
\end{tabular}}
\label{tab:res_human_perf_query}
\end{table}

\section{Interactive Refinement Ablation}
\label{sec:appendix_abl}
Full numbers for the interactive query refinement ablation discussed in the main
paper. The subset comprises the real-user queries of Scenario~02 that carry a spatial
(positional) constraint.

\begin{table}[h]
\caption{\textbf{Ablation of interactive query refinement} on real-user queries
with spatial constraints (Grounding-DINO backbone). Numbers are not directly
comparable to the full real-user set reported in the main paper.}
\centering
\resizebox{\columnwidth}{!}{
\begin{tabular}{l|clll}
\toprule
\textbf{Model Variant} & \textbf{HOTA} & \textbf{DetA} & \textbf{AssA} & \textbf{IDF1} \\ \toprule
\ourmodel (w/ Refinement)  & 47.4 & 38.9 & 59.6 & 69.3 \\ \midrule
\ourmodel (w/o Refinement) & 35.6 & 32.0 & 41.3 & 51.2 \\ \bottomrule
\end{tabular}}
\label{tab:ablation_refinement}
\end{table}

\section{Conventional MTMCT Comparison}
\label{sec:appendix_cityflow}
Tab.~\ref{tab:cityflow_results} reports \ourmodel on the CityFlow S02
(validation) set~\cite{tang2019cityflow} under the simplified query ``Track all
vehicles that appear in at least two cameras'', which removes language
conditioning from the evaluation. 
\reviewedit{It is included only as a conventional MTMCT reference point.}
\begin{table}[t]
\caption{\textbf{Performance on CityFlow} under the conventional MTMCT protocol,
using YOLOv5 and Grounding-DINO as representative backbones.}
\centering
\resizebox{\columnwidth}{!}{
\begin{tabular}{l|l|c|rrrr}
\toprule
\multicolumn{1}{c|}{\textbf{Model}} &
\multicolumn{1}{c|}{\textbf{Detection Backbone}} & \textbf{Training Source} &
\multicolumn{1}{c}{\textbf{HOTA}} &
\multicolumn{1}{c}{\textbf{DetA}} &
\multicolumn{1}{c}{\textbf{AssA}} &
\multicolumn{1}{c}{\textbf{IDF1}} \\ \toprule
Shim et al.~\cite{shim2023fast} & YOLOv7-E6 & MS COCO & 59.68 & 54.23 & 68.35 & 81.95 \\
TransRMOT \cite{wu2023referring} & Def.\ DETR & MS COCO & 28.82 & 23.35 & 36.65 & 29.38 \\
TempRMOT \cite{zhang2024bootstrapping} & Def.\ DETR & MS COCO & 24.94 & 16.35 & 39.34 & 24.74 \\ \midrule
\multirow{2}{*}{\textbf{\ourmodel (ours)}} & YOLOv5 & MS COCO & 62.99 & 58.35 & 69.59 & 86.84 \\
\cmidrule{2-7}
& Grounding-DINO & Open-vocab. & 40.27 & 48.88 & 33.41 & 42.30 \\
\bottomrule
\end{tabular}}
\label{tab:cityflow_results}
\end{table}

\section{Geometric and Temporal Pre-processing}
\label{sec:appendix_geo}
The base MTMCT algorithm requires overlap masks and temporal alignment across
cameras. Instead of relying on manual calibration, we construct both
automatically using the dataset-provided homography matrices $\mathbf{H}_i$.

\paragraph{Overlap masks.}
For each pixel $\mathbf{u}$ in camera $i$ we project it to world coordinates via
$\mathbf{H}_i^{-1}\mathbf{u}$ and test whether the projected point re-projects
into a valid in-ROI pixel of camera $j$; the pixels for which this holds define
the overlap region between the two views.

\paragraph{Temporal alignment.}
We define a canonical timeline $\tau$ and a mapping
$\text{temp\_align}[\textit{cam}][\tau]\rightarrow f$ from canonical time to raw
frames, estimated from GPS-matched foot-points of co-observed vehicles (matched
within $0.0003^{\circ}$, $\approx\!33\,\text{m}$). Depending on the number of
matched samples we apply (i)~isotonic regression for dense matches ($\geq 100$),
(ii)~a median offset for moderate matches (4--99), and (iii)~linear
interpolation from timestamp anchors otherwise. On CityFlow Scenario~02 the
automatically generated masks reach IoU $0.80$--$0.97$ against manual
annotations, with less than $0.1\%$ difference in the resulting IDF1.

\paragraph{Clustering threshold.}
The cross-camera association described in Sec.~4.4 of the main paper groups
tracklets by complete-linkage clustering with a threshold selected by the Dunn
index. The two thresholds mentioned in the main paper play different roles:
$\tau_{\mathrm{mtmc}}=0.65$ is an \emph{upper bound} on the search space, and
the operating threshold is the candidate linkage distance below that bound with
the largest jump in Dunn index. If no candidate falls below the bound, the
smallest linkage distance is used, which leaves each tracklet in its own
cluster. Pairs that violate the overlap constraint are excluded before
clustering by assigning them a distance far above the bound, so they can never be
merged regardless of appearance similarity.

\section{Agent Prompts and Latency}
\label{sec:appendix_prompts}
We describe how the two agents are prompted. The Requirement Analyzer prompt is reproduced verbatim below; the Task Manager prompt will be released with the code.
Query types T1--T6 in the prompt correspond to QF-1--QF-6 of the main paper.

The Requirement Analyzer acts as an analyst for a multi-camera tracking system.
It reads the user query and extracts the fields of the \emph{Requirement Schema}
defined in Sec.~4.1 of the main paper (color, vehicle type, camera ID, time
range, direction, zone, action, lane and relation), which is the JSON object
emitted in Step~3 of the prompt. It then tags whatever is missing, asks the user
targeted clarification questions, and returns a complete structured summary.

The Task Manager takes the resolved query, decomposes it into camera- and
time-specific sub-tasks, selects between the available detectors according to
the parsed requirements, and schedules execution. Both agents run at
temperature~0, and the system falls back to rule-based parsing when the model
output is malformed. The parsing model is OpenAI gpt-4o-mini; the full prompt text will be released with the code.

\paragraph{Latency.}
The mean end-to-end response time of the parsing model is $1.6$\,s per query,
ranging from $0.5$\,s to $3$\,s depending on query complexity and network
conditions; approximately 80\% of queries are parsed within $1$\,s. Parsing
therefore contributes a small fraction of the total processing time reported in
Tab.~\ref{tab:res_human_perf_query}.

\subsection*{Requirement Analyzer prompt}
A condensed version of the system prompt is reproduced below.

\begingroup\small
\noindent\textbf{Role.} You are an analyst agent for a multi-camera tracking
system. Your task is to receive queries from users and extract important
properties called components, then ask for missing information and provide
complete summaries.

\noindent\textbf{Query formats.} (T1) Basic: color; vehicle type; camera ID;
time range (start/end, seconds); additional camera IDs; dataset. (T2)
Directional: T1 + starting/ending position (left, right, top, bottom). (T3)
Zone-based: T2 + starting/ending zone (zones follow the format ``1a, 1b, 1c''
keyed to the camera ID). (T4) Action: T3 + action type (turns left, turns
right, goes straight, stops at traffic light). (T5) Relational: T4 + relation
type (follow, are followed by, move in front of, move in back of, drives
alongside, go behind) and object description. (T6) Lane-based: T5 + lane type
(horizontal/vertical), lane number (1--3), lane direction. (The prompt's lane
range is indicative of the initial camera set; higher lane indices, e.g.,
lane~4 in the QF-6 example of Sec.~\ref{sec:appendix_ai_examples}, are recorded
in the schema's numeric lane field.) Mixed: any combination; may miss components or contain additional information.

\noindent\textbf{Component vocabulary.} Colors: white, black, red, blue, gray,
grey, silver, brown, maroon, purple, green, tan, gold, yellow. Vehicle types:
sedan, SUV, truck, pickup, van, wagon, hatchback, coupe, jeep, minivan.
Datasets: \ourdata, CityFlow. Scenario
cameras (\ourdata/CityFlow): S1: cam 1--5; S2: cam 6--9; S3: cam 10--15; S4:
cam 16--40; S5: cam 10, 16--29, 33--36; S6: cam 41--46 (S6 exists in CityFlow but is
outside \ourdata, whose 40 cameras come from S1--S5).

\noindent\textbf{Response protocol.} Step 1, initial analysis: list all
components present in the query and tag partial information. Step 2, questions:
restate available information and ask targeted questions for missing fields
(dataset, vehicle type, color, initial camera, reappearance cameras, time
range, then lane/zone/direction/action details), iterating at most twice or
stopping when the user indicates there is no more information. Step 3, output:
emit the gathered information as a fixed JSON object (vehicle
\{color,\,type\}; camera \{main\_camera\_id,\,additional\_cameras\};
time\_range \{start,\,end\}; direction \{start,\,end\}; zone
\{start,\,end\}; action; relation; lane \{type,\,number,\,direction\};
dataset; detection\_model), with \texttt{null} for missing components and no
explanatory text.
\endgroup

\section{Qualitative Example}
\label{sec:appendix_qualitative}
\begin{figure*}[t]
\centering
\includegraphics[width=0.98\textwidth]{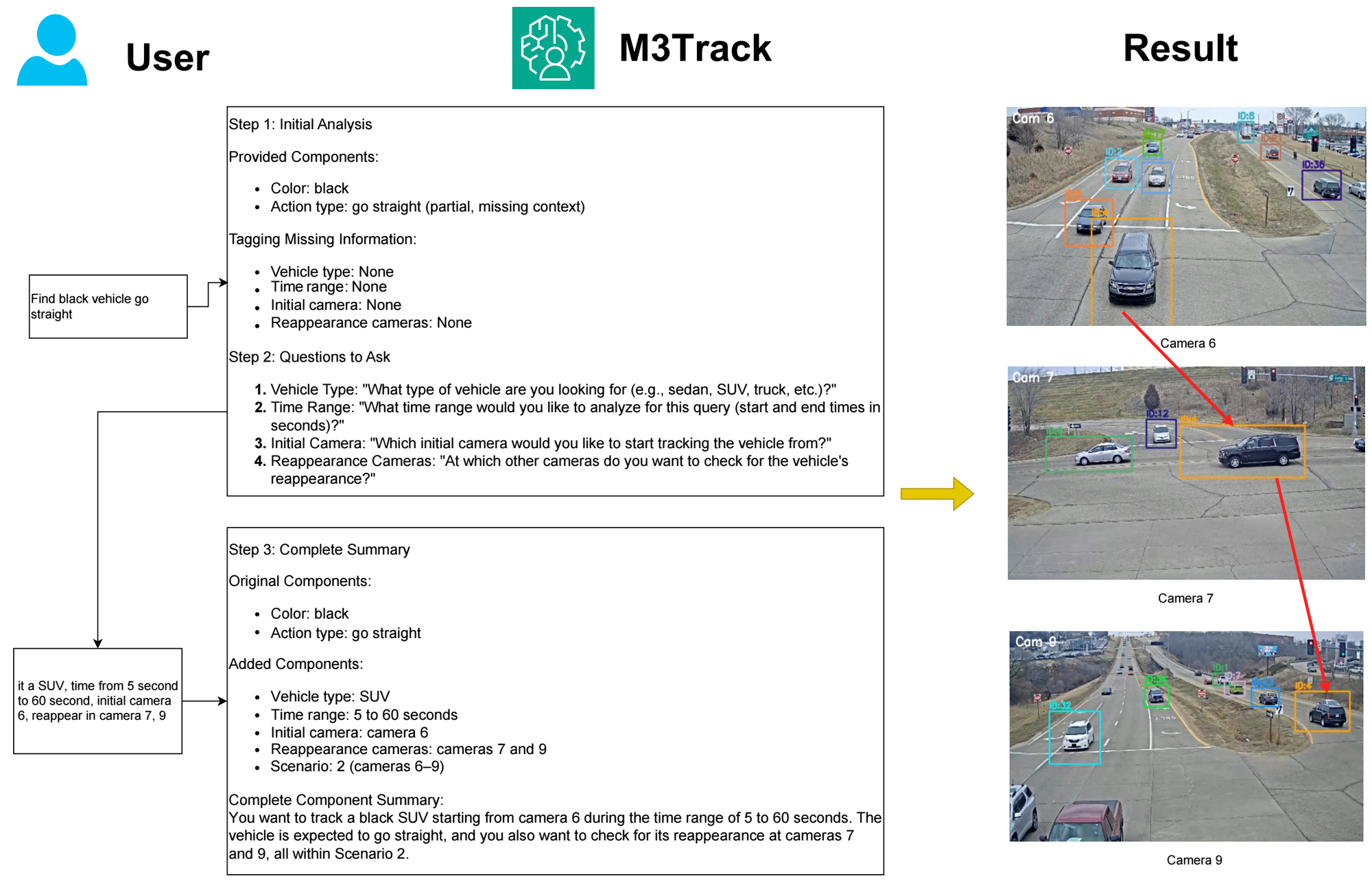}
\caption{\textbf{Interactive clarification and tracking on a real underspecified
query.} For the raw query ``Find black vehicle go straight'', the Requirement
Analyzer tags the missing fields (vehicle type, time range, cameras), asks four
targeted questions, and resolves the query to a black SUV starting from camera~6
and reappearing in cameras~7 and~9 within a 5--60\,s window; the pipeline then
tracks the target across the three cameras (right).}
\label{fig:qualitative}
\end{figure*}

Fig.~\ref{fig:qualitative} traces the full pipeline on a real underspecified
query, ``Find black vehicle go straight''. The Requirement Analyzer extracts the stated
fields, tags the missing ones, and asks four clarification questions;
the participant's answers resolve the query into the running example of the main
paper, a black SUV from camera~6 reappearing in cameras~7 and~9, which the
pipeline then tracks across all three cameras. Both the raw and the resolved forms are retained, which is what makes the
refinement ablation of Sec.~\ref{sec:appendix_abl} possible.

\section{Dataset Documentation}
\label{sec:datasheet}
We provide datasheet-style documentation following Gebru et al.~\cite{gebru2021datasheets}.

\paragraph{Motivation.}
\ourdata was created to enable the study of Referring Multi-Target Multi-Camera
Tracking (Refer-MTMCT): retrieving identity-consistent trajectories across a camera network from a free-form natural-language query. Existing referring tracking datasets are confined to a single camera and constrained query formats, and existing multi-camera datasets do not provide language grounded in cross-camera identity together with spatio-temporal predicates. \ourdata fills this gap.

\paragraph{Composition.}
The benchmark contains 96 videos ($\approx$3.25 hours, 116{,}000+ frames) from 40 roadside cameras, 666 global multi-camera trajectories, and 4{,}362 manually annotated trajectory-level descriptions. It provides 7{,}699 queries: 6{,}000 AI-generated (six query formats, 1{,}000 each) and 1{,}699 real-user queries. Each trajectory carries four attribute groups: appearance (color, vehicle type), motion (direction, action), spatial context (camera ID, zone, lane), and relational information, with per-camera zones labeled as entry, exit, and transition regions.

\paragraph{Collection process.}
\ourdata is built on the public CityFlow~\cite{tang2019cityflow} traffic
benchmark, which supplies the raw video and the global cross-camera identity annotations. Trajectory-level descriptions were produced by four annotators (a combined 960 hours) against a unified standard, cross-checked to resolve inconsistencies, and verified in a final 120-hour quality-control pass. AI-generated queries were synthesized from the ground-truth database and validated by executing each query against it; real-user queries were contributed by 50 participants with diverse backgrounds (students, computer scientists, transportation professionals, managers).

\paragraph{Uses.}
The dataset is intended for benchmarking language-driven multi-camera tracking. It contains vehicles rather than persons and is not intended for person surveillance or re-identification.

\paragraph{Institutional review.}
The study collects free-form text queries about vehicles from adult volunteers.
It records no personally identifying information, no images or audio of
participants, and no sensitive attributes, and it poses no more than minimal
risk. Participants were recruited by convenience sampling among students and
professional contacts of the authors; participation was voluntary and could be
discontinued at any time.
Participants were informed of the research purpose and of the fact that their
queries would be released publicly, and gave consent before contributing. No
compensation was offered, so participation carried no financial inducement. The
released queries describe vehicles and contain no information identifying the
participant. 

\paragraph{Distribution and license.}
\ourdata is released in three components. (i)~The \emph{annotations,
natural-language queries, and query-to-trajectory ground truth} are available at the repository
under CC~BY-NC~4.0. (ii)~The \emph{evaluation protocol, evaluator, agent prompts,
and reference implementation} will be released under the PolyForm Noncommercial
License~1.0.0. (iii)~The
\emph{source videos will not be redistributed}: they remain the property of the
AI City Challenge and must be obtained from it under its own terms, to which this
release is subordinate. This follows the practice of
CityFlow-NL~\cite{feng2021cityflow}, which distributes language annotations over
CityFlow without redistributing the imagery. All components are intended for research use; the annotation and query license does not permit commercial use, and none of the components may be used for surveillance of individuals.

\paragraph{Maintenance.}
The authors will maintain the benchmark after release. Issues and corrections can be reported through the project repository or to qlg645@mocs.utc.edu.

{\small
\bibliographystyle{ieeenat_fullname}
\bibliography{M3Track-ref}
}